\documentclass[lettersize,journal]{IEEEtran}
\usepackage{amsmath,amsfonts,amssymb}
\usepackage{algorithmic}
\usepackage{algorithm}
\usepackage{array}
\usepackage[caption=false,font=normalsize,labelfont=sf,textfont=sf]{subfig}
\usepackage{textcomp}
\usepackage{stfloats}
\usepackage{url}
\usepackage{verbatim}
\usepackage{graphicx}
\usepackage{cite}
\usepackage[table]{xcolor}
\newcommand{\revised}[1]{\textcolor{black}{#1}}

\begin{document}

\title{Cross-lingual Embedding Clustering for Hierarchical Softmax in Low-Resource Multilingual Speech Recognition}

\author{Zhengdong Yang, Qianying Liu, Sheng Li, Fei Cheng, Chenhui Chu
\thanks{\revised{A preliminary version was posted on arXiv \cite{yang2025crosslingualembeddingclusteringhierarchical}.
This revised manuscript includes additional experiments and analyses and is not identical to the preprint.}}}
        % <-this % stops a space
%\thanks{This paper was produced by the IEEE Publication Technology Group. They are in Piscataway, NJ.}% <-this % stops a space
%\thanks{Manuscript received April 19, 2021; revised August 16, 2021.}}

% The paper headers
%\markboth{Journal of \LaTeX\ Class Files,~Vol.~14, No.~8, August~2021}%
%{Shell \MakeLowercase{\textit{et al.}}: A Sample Article Using IEEEtran.cls for IEEE Journals}

%\IEEEpubid{0000--0000/00\$00.00~\copyright~2021 IEEE}
% Remember, if you use this you must call \IEEEpubidadjcol in the second
% column for its text to clear the IEEEpubid mark.

\maketitle

\begin{abstract}
We present a novel approach centered on the decoding stage of Automatic Speech Recognition (ASR) that enhances multilingual performance, especially for low-resource languages. 
It utilizes a cross-lingual embedding clustering method to construct a hierarchical Softmax (H-Softmax) decoder, which enables similar tokens across different languages to share similar decoder representations.  
It addresses the limitations of the previous Huffman-based H-Softmax method, which relied on shallow features in token similarity assessments.
Through experiments on a downsampled dataset of 15 languages, we demonstrate the effectiveness of our approach in improving low-resource multilingual ASR accuracy. 
\end{abstract}


\section{Introduction}

%TODO: First decide whether to extend the current paper or to make a new proposal, afterward decide the motivation etc.

The field of Automatic Speech Recognition (ASR) has seen significant advancements in recent years, which enables machines to generate high-quality transcriptions of human speech. However, the field still faces the challenge of covering only around 100 languages among approximately 7,000 languages spoken globally. Multilingual ASR models have emerged as a promising solution to this challenge, capable of supporting multiple languages within a single ASR framework and learning universal features that transfer knowledge from rich-resource to low-resource languages.

Early ASR systems relied on context-dependent deep neural network hidden Markov models~\cite{dnnhmm} that require language-specific pronunciation lexicons, which made it challenging to develop multilingual ASR systems. 
%Such models, however, demonstrated limitations when applied to low-resource languages due to insufficient exploration of modeling techniques tailored to these languages. 
The shift towards end-to-end (E2E) attention-based models~\cite{hinton2012deep,graves2014towards,bahdanau2016end} marked a significant turning point, offering simplified training processes and reducing the dependency on pronunciation lexicons, prompting increased research in the field of multilingual ASR.
Recent studies in E2E models have explored the learning of universal representations across different languages at the encoding stage, employing methods like decomposing phonemes into universal articulatory attributes~\cite{DBLP:conf/aaai/LiDM0BM20, li2019end, li2021hierarchical} and transfer learning between different languages with adapters~\cite{DBLP:journals/taslp/HouZWWQXS22}. Meanwhile, pre-trained self-supervised learning multilingual acoustic models~\cite{DBLP:conf/nips/BaevskiZMA20, DBLP:conf/nips/BaevskiHCA21, DBLP:journals/corr/abs-2111-02735, DBLP:journals/taslp/HsuBTLSM21, DBLP:journals/jstsp/ChenWCWLCLKYXWZ22} and multilingual speech corpora~\cite{DBLP:conf/lrec/WangPWG20, DBLP:conf/interspeech/WangWGP21, DBLP:conf/lrec/ArdilaBDKMHMSTW20, cmu-data, DBLP:conf/icassp/Black19} are investigated for learning pre-trained cross-lingual representations.

While various studies have been established for encoder representations, improvements in the decoding stage have long been overlooked. As multilingual models rapidly gain popularity, intuitively, there is an urgent need to explore how connections between similar tokens across different languages can be enhanced during decoding. 
A recent study~\cite{liu2023hierarchical} employs hierarchical Softmax (H-Softmax)~\cite{DBLP:conf/aistats/MorinB05}, an approximation Softmax inspired by binary trees, to make similar tokens across different languages share similar decoder representations, thereby enhancing model performance. 
%This approach also simplifies the Softmax operation into more manageable binary classifications
This approach also breaks down the computationally expensive Softmax function into multiple binary classification steps, boosting model efficiency~\cite{mohammed2018effectiveness}. However, we recognize one major limitation in this study, where such a Huffman-based approach relies on shallow features, i.e., token frequency, for token similarity assessment, which may not fully capture the nuances of cross-lingual correlations.

%Building upon this, we explore two methods for constructing the binary trees essential for H-Softmax: Huffman-based and embedding-based clustering. Initially, we adopt the Huffman coding to capture token similarity, motivated by linguistic studies that reveal distributional similarities among neighboring languages~\cite{artetxe-etal-2018-robust}. However, we recognize the limitations of the Huffman-based approach, particularly its reliance on shallow features, i.e. token frequency, for token similarity assessment, which may not fully capture the nuances of cross-lingual correlations. 

To overcome the limitation, we propose an embedding-based method inspired by the success of cross-lingual embeddings in natural language processing (NLP). 
These embeddings are typically generated by models trained on massive, diverse datasets that include text from multiple languages, using techniques such as masked language modeling and translation language modeling to capture semantic equivalences across languages~\cite{conneau2019cross,feng2022language}. They effectively model linguistic relations, facilitating significant improvements in tasks such as machine translation and multilingual text classification. 
Building on this foundational success, our method employs hierarchical clustering of cross-lingual embeddings to construct the hierarchical decoder, which allows for a better understanding of token similarity.
The main contributions of this paper are:
\begin{itemize}
    \item We introduce an embedding-based H-Softmax decoding strategy to enhance the capability of multilingual ASR systems. Our experiments demonstrate the superiority of our method over the Huffman-based baseline in a low-resource setting.
    This method is notably scalable, as it can easily incorporate increasingly powerful embedding models that may emerge in the future to further enhance its effectiveness.
    \item We investigate different approaches of hierarchical clustering of embeddings for automatically capturing cross-language token similarities for low-resource multilingual ASR.
\end{itemize}

The remainder of this paper is structured as follows. 
Section II surveys related works on multilingual ASR models, decoding strategies in ASR, and cross-lingual embeddings. 
Section III introduces the preliminary framework that our proposed method is based on, including Huffman-based tree construction and the H-Softmax decoder.
Section IV details our methodology of embedding-based tree construction, which includes obtaining the cross-lingual embeddings and hierarchical clustering of embeddings.
Section V presents the experimental results obtained from evaluating our approach on a downsampled dataset comprising 15 languages, demonstrating our method's effectiveness in enhancing the accuracy of low-resource multilingual ASR.
In Section VI, we conduct several comprehensive analyses to understand our method's performance further. 
Finally, Section \revised{VII} concludes the paper by summarizing our findings and contributions.

\section{Related Work}

%TODO: Collection Related Work and write the section
%Reference: Language-Universal Phonetic Representation in Multilingual Speech Pretraining for Low-Resource Speech Recognition https://arxiv.org/abs/2305.11576

\subsection{Multilingual ASR}

Recent advancements in multilingual ASR have featured different methods to address the challenges associated with low-resource languages. Research has demonstrated that zero-shot learning can be effectively applied to phoneme recognition, as evidenced by a study where unseen phonemes were recognized by decomposing them into articulatory attributes~\cite{DBLP:conf/aaai/LiDM0BM20}). Cross-lingual speech adaptation using adapter modules has shown promise in improving the efficiency of transfer learning with significantly fewer trainable parameters while also reducing word error rates~\cite{DBLP:journals/taslp/HouZWWQXS22}.

Pre-trained self-supervised learning models also enhance ASR performances in multilingual settings. Wav2Vec~\cite{DBLP:conf/nips/BaevskiZMA20} demonstrates the potential of self-supervised learning by using masked prediction techniques in the latent space, achieving remarkable results with very limited labeled data. Following this, HuBERT~\cite{DBLP:journals/taslp/HsuBTLSM21} introduces an innovative approach by utilizing offline clustering to provide targets for a BERT-like prediction loss, focusing on masked regions only. Additionally, WavLM~\cite{DBLP:journals/jstsp/ChenWCWLCLKYXWZ22} has extended these benefits to a broader range of speech-processing tasks by incorporating novel techniques such as masked speech prediction combined with speech denoising, thereby improving the model's effectiveness across various speech tasks.

In parallel, the development of large multilingual speech corpora has been crucial to advancements in multilingual ASR. 
The Common Voice project~\cite{DBLP:conf/lrec/ArdilaBDKMHMSTW20} has compiled a vast multilingual corpus that supports ASR research across multiple languages.
Likewise, the CMU Wilderness Multilingual Speech Dataset~\cite{DBLP:conf/icassp/Black19} offers a substantial resource with over 700 different languages, providing audio, aligned text, and word pronunciations for each language.

Despite these advancements, previous research has mainly focused on enhancing encoder representations, leaving a significant gap in understanding and improving decoding mechanisms for multilingual ASR. 

\subsection{Hierarchical Softmax}

Traditional Softmax, typically employed in the final layer of neural networks to produce a probability distribution over potential output classes, becomes computationally burdensome as the number of classes increases. H-Softmax, which organizes the output classes into a tree-based hierarchy, has emerged as a prominent solution to address the computational challenge. 

%Advancements in embedding models have included methods like negative sampling as an alternative to Softmax, aimed at improving training speed and quality of vector representations in contexts with large vocabularies~\cite{mikolov2013distributed}. 

Morin and Bengio~\cite{DBLP:conf/aistats/MorinB05} pioneered the exploration of using hierarchical output probability in neural network language models, showing that it can substantially reduce the computational cost without a significant loss in accuracy. 
Mnih and Hinton~\cite{DBLP:conf/nips/MnihH08} extended the hierarchical structure to the log-bilinear model, which consists of one linear hidden layer and a softmax output layer. 
Mikolov et al.~\cite{DBLP:journals/corr/abs-1301-3781} refined the definition of H-Softmax, incorporating it to accelerate the training of word embeddings.

%Additionally, an adaptive Softmax approach has been proposed, which significantly enhances training efficiency on contemporary hardware architectures by exploiting the unbalanced distribution of word frequencies and clustering them to optimize computational expectations~\cite{DBLP:conf/icml/GraveJCGJ17}.

Additionally, the capability of H-Softmax when involving a large number of classes has been specifically investigated. Research utilizing large-scale datasets has shown that H-Softmax may exhibit performance trade-offs as the number of classes increases. While it considerably improves efficiency and reduces computational demand, it can lead to degradation in classification accuracy when dealing with an extensive number of classes~\cite{mohammed2018effectiveness}.

Despite the proven benefits of H-Softmax in NLP, its direct utilization within the domain of ASR, particularly in low-resource and multilingual contexts, has not been thoroughly explored.

\subsection{Cross-Lingual Embeddings}

Research on cross-lingual embeddings has underscored their potential to bridge linguistic gaps and enable the sharing of semantic spaces across languages. 

One notable work in this area is the cross-lingual language model (XLM)~\cite{conneau2019cross}. It leverages the transformer architecture with different language modeling objectives to learn a shared representation for multiple languages simultaneously. 
XLM has demonstrated substantial improvements in tasks including cross-lingual classification, machine translation, low-resource language modeling, and unsupervised cross-lingual embeddings

Additionally, a BERT-based embedding model, Language-agnostic BERT Sentence Embedding (LaBSE)~\cite{feng2022language}, has also been proposed. LABSE utilizes a combination of language model pre-training and dual encoder translation ranking to effectively capture semantic meanings across languages, making it highly effective for tasks such as semantic search, text classification, and information retrieval in multilingual settings.  

%In summary, our work intersects with and extends upon several key areas of ASR research. By focusing on the decoding process and introducing a novel embedding-based hierarchical decoder, we contribute to the ongoing dialogue on enhancing multilingual ASR systems, particularly for languages that have traditionally been underrepresented in speech technology.

\section{Preliminary}

\begin{figure*}[htb]
  \setlength{\abovecaptionskip}{15pt}
  \centering
  \includegraphics[width=0.9\textwidth]{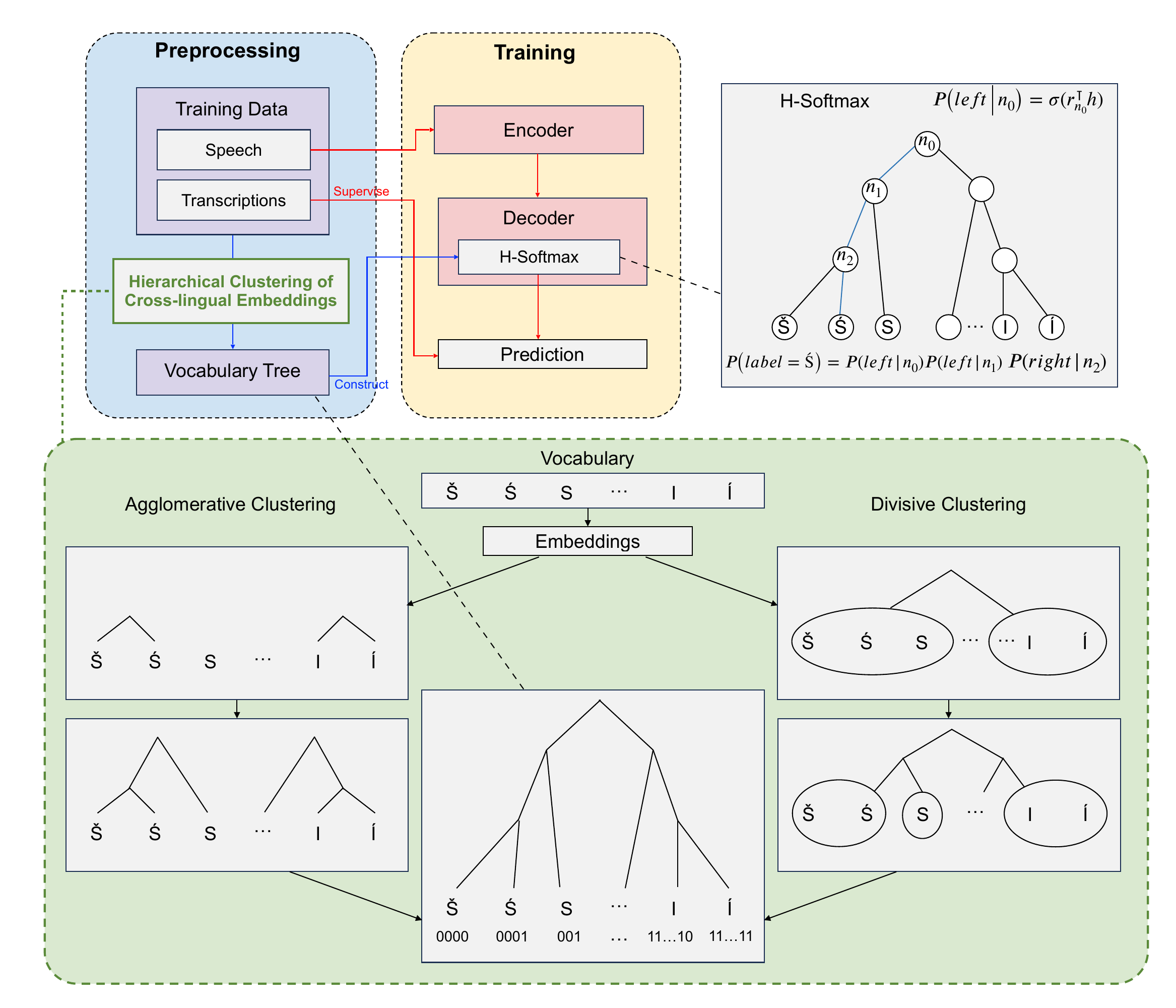}
  \caption{The flowchart of our framework for ASR with H-Softmax. The blue line represents how the H-Softmax network is determined, and the red line represents how the ASR model is trained. The green area at the bottom shows the detail of the proposed hierarchical clustering of cross-lingual embeddings. }
  \label{fig:proposed}
\end{figure*}

We first introduce the framework of \cite{liu2023hierarchical}, which investigated employing H-Softmax for the decoding stage of ASR, with Huffman coding adopted for the tree construction. The upper part of Figure \ref{fig:proposed} illustrates the general structure of this framework, with Huffman coding replaced by the proposed method that we will introduce in the next section.
In this section, we introduce this preliminary work that mainly includes two parts: 

i. A binary tree consisting of tokens in the vocabulary (hereinafter referred to as \emph{vocabulary tree}) is created based on Huffman coding.

ii. An H-Softmax decoder is constructed to leverage the vocabulary tree for ASR training and inference. 

\subsection{Huffman-based Vocabulary Tree}

Based on the assumption that neighboring languages exhibit similar token distributions, Huffman coding is applied to all the tokens in the vocabulary for constructing the vocabulary tree. Consider a multilingual token vocabulary $V=\{t_1, t_2, ...t_N\}$, where $N$ represents the total number of unique tokens in the vocabulary. A term frequency set $S_p=\{p_{t_i}\}_{i=1}^N$ is compiled, where $p_{t_i}$ denotes the frequency of token $t_i$ across all languages in the training set, treating identical tokens from different languages as a single entity.

Utilizing $S_p$, a Huffman tree for $V=\{t_i\}$ is constructed through a process of frequency clustering. This involves initially treating each token $t_i$ as an isolated node with a weight corresponding to its frequency $p_{t_i}$. These nodes are then progressively merged in a bottom-up manner according to their frequencies. Specifically, in each iteration, the two nodes with the lowest frequencies are selected and combined into a new node, where the new node's frequency is the sum of the two selected nodes' frequencies. The process repeats until a single node remains, serving as the root of the Huffman tree. 

The Huffman tree construction method ensures that the most frequent tokens are placed closer to the root, enabling an efficient binary encoding scheme. For our study specifically, this approach allows tokens with similar frequencies to be positioned at comparable levels within the tree's hierarchy, effectively capturing token similarity across different languages. However, this Huffman-based method only utilized shallow features, which could struggle to capture cross-lingual correlations effectively and limit the performance. 

\subsection{Hierarchical Softmax Decoder}

%NOTE: Mainly copied from ICASSP paper
%We first generate the Huffman code of each token.
%further recursively generate the Huffman code by assigning 0 to the left subtree and 1 to the right subtree. 

The vanilla Softmax, utilized by previous sequence generative models for ASR to predict the next token, is replaced with H-Softmax in this framework. 
H-Softmax organizes the output vocabulary into a tree where the leaves are the vocabulary tokens, and the intermediate nodes are latent variables \cite{DBLP:conf/aistats/MorinB05}. 

The tree generated in previous subsections is used as the tree for H-Softmax. 
Specifically, the binary tree H-Softmax described in \cite{mikolov2013distributed} is adopted. 
The decoding procedure is transformed into predicting one leaf node of the binary tree at each timestep. 
Each leaf node, representing a token, could be reached by a path from the root through the inner nodes. 
Given the transformer output hidden state $h$ and trainable node representation vectors $\{r_i\}$, the final possibility of a leaf node $w_i$ could be represented as:

% \begin{align*}
%         P(label = {leaf}) &= \prod^{parent({leaf})}_{root} P(n_k, {path}_k) \\&= \prod^{parent({leaf})}_{root} \sigma (v_{n_i}, h)
% \label{eq:prob}
% \end{align*}
\vspace{-12pt}
\begin{equation}
\begin{aligned}
        P(label = {w}_i) = \prod^{{path}} P({path}_k|n_k) \\
        = \prod^{{path}}
        % \sigma (r_k^\intercal h)
        \begin{cases}
        \sigma (r_k^\intercal h) & \text{if  } {path}_k = left, \\
        1-\sigma (r_k^\intercal h) & \text{if  }  {path}_k = right. 
        \end{cases}
\end{aligned}
\label{eq:prob}
\end{equation}

\noindent where $n_k$ stands for the $k$th node on the path from the root to ${w}_i$ such as $n_0$, $n_1$ and $n_2$ in Figure~\ref{fig:proposed}; ${path}_k$ stands for the branch leading towards ${w}_i$ which is $left$ or $right$, and $\sigma(\cdot)$ stands for sigmoid function. 

\begin{figure}[ht]
  % \centering
    \vspace{-10pt}   % 本文と図の間隔微調整用
     \includegraphics[width=1\columnwidth]{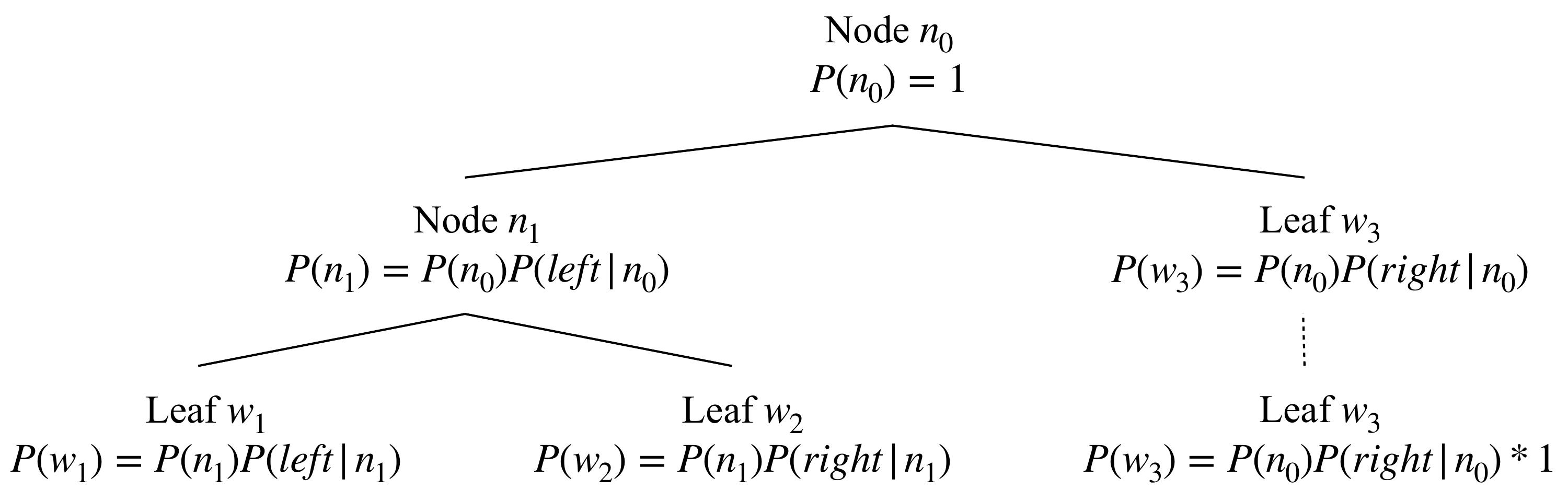}
    \vspace{-20pt}     % 図とキャプションの間隔微調整用
    \caption{A typical H-Softmax tree structure. Leaf $w_3$ has a virtual child with the same probability of aligning each leaf node to the same depth, so it is conceptually possible for path vectorization.}
    \label{fig:tree}
    \vspace{-5pt}   % キャプションと本文の間隔微調整用
\end{figure}

By decomposing the Softmax to a binary tree, H-Softmax reduces the decoding time complexity from O(V) to O(log(V)), and the train time complexity remains O(Vlog(V)). 
%revious H-Softmax implementation\cite{mohammed2018effectiveness} is on CPU, considering the order of magnitude difference between CPU's and GPU's FLOPs, the challenge of improving the efficiency of model training lies in designing an implementation of H-Softmax for GPU training. 
Furthermore, a vectorization algorithm is used to accelerate training and decoding procedures on GPU. 
Consider a typical H-Softmax tree structure shown in Fig~\ref{fig:tree}. Then, log-probability calculations from Eq~\ref{eq:prob} can be vectorized to the followings:
\begin{align*}
&\begin{bmatrix}
\log P(w_1)\\ 
\log P(w_2)\\ 
\log P(w_3)
\end{bmatrix}
% = \begin{bmatrix}
% \log[P(left|n_0)P(left|n_1)]\\ 
% \log[P(left|n_0)P(right|n_1)]\\ 
% \log[P(right|n_0)]
% \end{bmatrix} 
= \begin{bmatrix}
\log[\sigma(r_1^\intercal h)\sigma(r_2^\intercal h)]\\ 
\log[\sigma(r_1^\intercal h)(1-\sigma(r_2^\intercal h))]\\ 
\log[1-\sigma(r_1^\intercal h)]
\end{bmatrix} \\
% &=\begin{bmatrix}
% \log\sigma(r_1^\intercal h) + \log\sigma(r_2^\intercal h)\\ 
% \log\sigma(r_1^\intercal h) + \log(1-\sigma(r_2^\intercal h))\\ 
% \log(1-\sigma(r_1^\intercal h)) + \log 1
% \end{bmatrix} \\
&=\sum\limits_{column}\log\begin{bmatrix}
1*\sigma(r_1^\intercal h)+0 & 1*\sigma(r_2^\intercal h)+0\\ 
1*\sigma(r_1^\intercal h)+0 & -1*\sigma(r_2^\intercal h)+1\\ 
-1*\sigma(r_1^\intercal h)+1 & 0*\sigma(r_2^\intercal h)+1
\end{bmatrix} \\
% &=\sum\limits_{column}\log\left(
% \begin{bmatrix}
% 1 & 1\\ 
% 1 & -1\\ 
% -1 & 0
% \end{bmatrix}
% \circ
% \begin{bmatrix}
% \sigma(r_1^\intercal h) & \sigma(r_2^\intercal h)\\ 
% \sigma(r_1^\intercal h) & \sigma(r_2^\intercal h)\\ 
% \sigma(r_1^\intercal h) & \sigma(r_2^\intercal h)
% \end{bmatrix}
% +
% \begin{bmatrix}
% 0 & 0\\ 
% 0 & 1\\ 
% 1 & 1
% \end{bmatrix}
% \right) \\
&=\sum\limits_{column} \log(Sign \circ \sigma(p) + Bias)
\label{eq:log_prob}
\end{align*}
where $Sign$ is a $3 \times 2$ matrix of the signs of $\sigma$, $Bias$ is a $3 \times 2$ matrix of the biases of $\sigma$ and $p$ is the result vector of the inner product between the node vectors and $h$.
After building the Huffman tree, the $Sign$ and $Bias$ matrices are fixed. 
So, in the training stage, leaf node log probabilities can only be acquired through vector operations. 

%For decoding, we only need leaves with the highest probabilities. 
%To directly calculate this objective, different from training, we also develop a path-encoding-based multi-layer beam searching on GPU for H-Softmax\footref{github} to retain the time efficiency advantage of time-space complexity O(log(V)) compared to vanilla Softmax's O(V). 

\section{Cross-lingual Embedding Clustering}

%Our proposed methodology for enhancing ASR in low-resource multilingual contexts employs a novel application of H-Softmax. 
Based on the framework introduced in the previous section, we propose a novel embedding-based approach to construct the vocabulary tree, as shown in the green area of Figure \ref{fig:proposed}.
Our proposed method can be divided into two parts: 

i. We obtain the cross-lingual embeddings for all the tokens in the vocabulary.

ii. We create the vocabulary tree by conducting hierarchical clustering to the obtained cross-lingual embeddings.

%iii. We construct an H-Softmax decoder that leverages the vocabulary tree for training and decoding. 

Below, we will introduce each part in detail.

%\subsection{Embedding-based Vocabulary Tree}

%Given that the Huffman-based approach to vocabulary tree generation relies solely on the superficial feature of token frequency, we propose an alternative embedding-based vocabulary tree method that utilizes more information-dense features. This method unfolds in two stages:

%i. Obtaining cross-lingual embeddings for the vocabulary.

%ii. Generating a binary tree from these embeddings using hierarchical clustering techniques.

%Next, we will provide an in-depth discussion of these two stages.

\subsection{Cross-Lingual Embedding}
%other embeddings
%our embedding which only needs monolingual data
%TODO: for the same token across languages, use the mean of all languages or use separate tokens for each language

%There are many pre-trained cross-lingual embeddings we can use for our embedding-based vocabulary tree construction. We tried following 2 : XLM and LABSE. Because in some case the low-resource language may not be supported by pre-trained embeddings, we also investigate using embeddings generated only with the our own training data. We adopt the method for cross-lingual mapping of monolingual embeddings in Artetxe’s work~\cite{artetxe-etal-2018-robust} as another way to obtain the embeddings. We only use the initialization part as the complete process only works for a bilingual setting.
%For the pre-trained cross-lingual embeddings: XLM~\cite{conneau2019cross} ... and LABSE~\cite{feng2022language}... 

\revised{Although it seems more reasonable to utilize pronunciation-based embeddings for ASR, we choose the semantic-based embeddings for constructing the vocabulary tree instead. The reasons include: (i) Mapping of text and pronunciation depends on G2P tools, but many of the low-resource languages lack reliable G2P or phoneme lexica. 
(ii) While the semantic-based embeddings are derived from text, at the character level, they mostly reflect orthography/phonotactics rather than high-level semantics, and thus are compatible with ASR objectives.}

\revised{We investigate} two different approaches to obtain the cross-lingual embedding: Directly using pre-trained cross-lingual embeddings, or conducting cross-lingual mapping of monolingual embeddings (hereinafter referred to as \emph{Mono-Map}). 

Although pre-trained cross-lingual embeddings are trained using large-scale data and are generally of higher quality, they may not always support a specific low-resource language.
In contrast, Mono-Map allows us to generate embeddings using exclusively our training data. 
%Next, we will introduce the details of these approaches.

\subsubsection{Pre-trained Cross-Lingual Embeddings}

We utilize various pre-trained cross-lingual embeddings for vocabulary tree construction. We have specifically experimented with two such embeddings introduced in Section II: XLM and LABSE.

XLM is based on a Transformer architecture, trained using a combination of different types of training approaches: Causal language modeling and masked language modeling on monolingual data in different languages, and translation language modeling on parallel data in different language pairs.
We investigated the effect of the model size with three different sizes of XLM: base, large, and XL.

LaBSE, on the other hand, is using an encoder-only BERT-based architecture. The model is first pre-trained using masked language modeling and translation language modeling. Then it is trained using dual-encoder translation ranking that optimizes for bilingual sentence similarity over translation pairs across languages. 

\subsubsection{Mono-Map}

We employ a strategy proposed by Artetxe et al.~\cite{artetxe-etal-2018-robust}. For our study, we have adapted only the initialization phase of their method, as the entire process is tailored for a bilingual setting and our application requires embedding mapping among multiple languages.

%The process starts with generating monolingual embeddings of the same length. For a certain language’s vocabulary, given the axes of the original monolingual embeddings $X$,  the similarity matrices $M_X = XX^T$. Sort the values in each row of $M_X$ to get $sorted(M_X)$. Use $sorted(\sqrt{M_X})$ as its magnitude is closer to the original embeddings. For the same token across languages, use the mean embedding of all languages.

The procedure starts by producing monolingual embeddings with uniform dimensionality. For the vocabulary for a specific language, we begin with the original monolingual embedding matrix $X$. 
We then compute the similarity matrices $M_X = XX^T$, where each row represents a set of similarities between one token and all the other tokens in the vocabulary.
Next, we take the square root to obtain $\sqrt{M_X}$, as this brings the values closer in magnitude to the original embeddings.
Finally, we sort the values within each row of $\sqrt{M_X}$ in ascending order, resulting in $sorted(\sqrt{M_X})$ as the mapped embedding matrix.

When dealing with tokens that appear in multiple languages' vocabulary, we utilize an average embedding strategy, taking the mean of the embeddings from all the languages for the shared token. This ensures the vocabulary to be consistent with that in the Huffman-based method or using pre-trained embeddings. 

\subsection{Hierarchical Clustering}

%With the cross-lingual embeddings obtained from the last section, we can conduct hierarchical clustering of the embeddings to construct the vocabulary tree. Hierarchical clustering here seeks to build a hierarchy of clusters based on the similarities among embeddings. Each cluster consists of two sub-clusters in the lower hierarchy, which corresponds to the structure of the binary tree. It has two main approaches: agglomerative and divisive.
%Agglomerative: a bottom-up approach where each embedding starts in its own cluster, and pairs of clusters are merged as one moves up the hierarchy.
%Divisive: a top-down approach where all embeddings start in one cluster, and splits are performed recursively as one moves down the hierarchy.
%Many variations are investigated, including different methods of clustering for two main approaches as well as different metrics for evaluating the similarities between 2 embeddings.

%We adopt the agglomeration strategy for hierarchical clustering that can be described as follows: i. Start with each embedding as its own cluster. ii. Merge the two clusters with the shortest distance into a single larger cluster. The distance between two clusters is the shortest distance between two embeddings in each cluster. iii. Repeat step ii until all embeddings belong to a single cluster. 

Having obtained cross-lingual embeddings, we advance to the hierarchical clustering of these embeddings to form our vocabulary tree. Hierarchical clustering in our context is applied to group the embeddings based on their mutual similarities, thereby forming a binary tree structure where each node is a cluster composed of two sub-clusters from the preceding level.

Given a set of $n$ token embeddings $E = \{e_0, e_1, ..., e_n\}$, the hierarchical clustering can be executed via two different approaches: 

\begin{itemize}
\item Agglomerative: A bottom-up approach starting with each embedding as a separate cluster $C_k = \{e_k\}$ and merging them sequentially. At each step, two clusters $C_i, C_j$ from the current cluster set $S = \{C_0, C_1, ..., C_{m}\}$ are combined to form a new parent cluster $C_p$, continuing until all embeddings are grouped into a single cluster.
\item Divisive: A top-down approach beginning with all embeddings in a single cluster $C_E = E$ and then splitting them iteratively. At each step, any cluster in the current cluster set $S = \{C_0, C_1, ..., C_{m}\}$ with more than one element is divided into two child clusters, continuing until each embedding is isolated in its own cluster.
\end{itemize}

\textbf{Agglomerative clustering} is characterized by its strategy of merging clusters. 
We explore several methods to select the next two clusters $C_i, C_j$ to be combined. 
%and the distance between two clusters denoted by $D$,
Different methods can be described as follows:

\noindent (Hereinafter, embeddings in different clusters will be denoted by $\mathbf{a}$, $\mathbf{b}$. The distance between two embeddings will be denoted by $d$.)

\begin{itemize}
\item Average: Merges clusters with the minimized average distance between all pairs of embeddings in the two clusters
\begin{equation}
    D_\text{A}(C_i, C_j) = \frac{1}{|C_i||C_j|} \sum_{\mathbf{a} \in C_i, \mathbf{b} \in C_j} d(\mathbf{a}, \mathbf{b})
\end{equation}
\item Weighted: Similar to Average clustering, but the distance is weighted in a way that two branches of a cluster have equal weights. The minimized distance can be calculated by recursively applying 
\begin{equation}
    D_\text{W}(C_i, C_j) = \frac{1}{2} [D_\text{W}(C_k, C_j) + D_\text{W}(C_l, C_j)]
\end{equation}
where $C_k$ and $C_l$ are the child clusters of $C_i$.

When there is only one embedding in each cluster, the distance between clusters is the distance between embeddings
\begin{equation}
    D_\text{W}(C_i, C_j) = d(\mathbf{a}, \mathbf{b}) \text{ if } C_i = \{\mathbf{a}\}, C_j = \{\mathbf{b}\} \\
\end{equation}
\item Centroid: Merges clusters with the minimized distance between the centroids of the clusters
\begin{equation}
    D_\text{C}(C_i, C_j) = \| \text{cent}(C_i) - \text{cent}(C_j) \|
\end{equation}
where 
\begin{equation}
\text{cent}(C) = \frac{1}{|C|} \sum_{\mathbf{a} \in C} \mathbf{a}
\end{equation}
\item Median: Merges clusters with the minimized distance between the medians of the clusters
\begin{equation}
    D_\text{M}(C_i, C_j) = \| \text{median}(C_i) - \text{median}(C_j) \|
\end{equation}
where ``median'' refers to the results of recursively calculating the median of two child clusters $C_k, C_l$ for the parent cluster $C_i$
\begin{equation}
    \text{median}(C_i) = \frac{1}{2} [\text{median}(C_k) + \text{median}(C_l)]
\end{equation}
\begin{equation}
    \text{median}(C_i) = \mathbf{a} \text{ if } C_i = \{\mathbf{a}\} \\
\end{equation}
In other words, Median is a weighted version of Centroid. 
\item Ward: Merges clusters that minimize the increase in within-cluster variance 
\begin{equation}
    \Delta \text{Var}(C_i, C_j) = \frac{|C_i||C_j|}{|C_i| + |C_j|} D_\text{C}(C_i, C_j)^2
\end{equation}
\end{itemize}

%More specifically, each method selects the next pair of clusters to be combined with different minimizations:
%$C_i, C_j, C_k, C_l$: These denote different clusters involved in the clustering operations. $C_i$ and $C_j$ are often the clusters being merged or compared, while $C_k$ and $C_l$ may represent newly formed clusters or other clusters involved in comparison or merging operations.

Conversely, \textbf{divisive clustering} is characterized by its systematic division of clusters. It is performed by repeating one of the following algorithms that divide a parent cluster $C_p$ into two child clusters $C_i, C_j$:

\begin{itemize}
\item 2-means: A special case of $k$-means algorithm where $k = 2$. It minimizes the sum of squared Euclidean distances between embeddings and their assigned cluster’s centroid
\begin{equation}
    J_\text{Means} =  \sum_{\mathbf{a} \in C_i} \| \mathbf{a} - \text{cent}(C_i) \|^2 +  \sum_{\mathbf{b} \in C_j} \| \mathbf{b} - \text{cent}(C_j) \|^2
\end{equation}
\item Spherical 2-means: Similar to 2-means, but uses cosine distance instead of squared Euclidean distance. The minimization objective is
\begin{equation}
    J_\text{Spherical} = \sum_{\mathbf{a} \in C_i} \frac{\mathbf{a} \cdot \text{cent}(C_i)}{\|\mathbf{a}\|\|\text{cent}(C_i)\|}
        + \sum_{\mathbf{b} \in C_j} \frac{\mathbf{b} \cdot \text{cent}(C_j)}{\|\mathbf{b}\|\|\text{cent}(C_j)\|}
\end{equation}
It can be seen as a version of 2-means where embeddings and centroids are projected to the surface of a high-dimensional sphere.
\item 2-medoids: Instead of centroid, it uses medoid, a selected embedding with the minimized total distances to other embeddings within the cluster.  
The algorithm minimizes the sum of these total distances of two clusters after division
%dissimilarities can be measured with different metrics.
\begin{equation}
    J_\text{Medoids} =  \sum_{\mathbf{a} \in C_i} d(\mathbf{a}, \mathbf{m}_i) +  \sum_{\mathbf{b} \in C_j} d(\mathbf{b}, \mathbf{m}_j)
\end{equation}
where
\begin{equation}
    \mathbf{m}_k = \underset{\mathbf{a} \in C_k}{\arg\min} \sum_{\mathbf{b} \in C_k} d(\mathbf{a}, \mathbf{b})
\end{equation}
\end{itemize}

For both clustering approaches, we examine various metrics to measure the similarity between embeddings, each offering a different perspective of the ``distance'' of two $m$-dimensional embeddings $\mathbf{a}$ and $\mathbf{b}$:

\begin{itemize}
\item Euclidean distance: Measures the ``straight-line'' distance between two embeddings in multidimensional space
\begin{equation}
    d_\text{Euc}(\mathbf{a}, \mathbf{b}) = \sqrt{\sum_{i=1}^{m} (a_i - b_i)^2}
\end{equation}

\item Standardized Euclidean distance: Similar to Euclidean distance, but each dimension is scaled by a weight
\begin{equation}
    d_\text{S-Euc}(\mathbf{a}, \mathbf{b}) = \sqrt{\sum_{i=1}^{m} \left( \frac{a_i - b_i}{\sigma_i} \right)^2}
\end{equation}
where $\sigma_i$ is the standard deviation of the $i$-th dimension across all embeddings in $E$.

\item City block (L1) distance: Computes the distance of embeddings by summing the absolute differences of their different dimensions
\begin{equation}
    d_\text{CB}(\mathbf{a}, \mathbf{b}) = \sum_{i=1}^{m} |a_i - b_i|
\end{equation}
\item Cosine distance: Assesses the cosine of the angle between two embeddings
%useful for understanding orientation, regardless of magnitude.
\begin{equation}
    d_\text{Cos}(\mathbf{a}, \mathbf{b}) = 1 - \frac{\mathbf{a} \cdot \mathbf{b}}{\| \mathbf{a} \| \| \mathbf{b} \|}
\end{equation}
\item Correlation distance: Distance based on Pearson correlation coefficient, with a focus on the degree to which they are linearly related
\begin{equation}
    d_\text{Cor}(\mathbf{a}, \mathbf{b}) = 1 - \frac{\sum_{i=1}^{m} (a_i - \bar{a})(b_i - \bar{b})}{\sqrt{\sum_{i=1}^{m} (a_i - \bar{a})^2 \sum_{i=1}^{m} (b_i - \bar{b})^2}}
\end{equation}
\end{itemize}

By applying these clustering methods and metrics, we aim to optimize the structure of the vocabulary tree to reflect the true linguistic relationships among the cross-lingual embeddings.

\section{Experiments}

\subsection{Dataset}

%downsampling method

We followed the setting of \cite{liu2023hierarchical} and sampled our speech-to-text datasets from Common Voice Corpus 11.0.
To investigate the impact of language relatedness on multilingual ASR performance with different methods, we selected three different linguistic groups: Romance, Slavic, and Turkic. For each group, we selected five languages from the corpus and constructed training, validation, and test sets using the validated data\footnote{Common Voice Corpus consists of voice recordings contributed by individuals, the validated data refers to those that have been reviewed and approved by community members for quality and accuracy.} of each language.

%First, we constructed the test sets with 10\% of the validated data. 
%To ensure efficient inference time, we limited the size of the test set so that it contained no more than 10,000 utterances for each language.

First, we constructed the test set for each language with a maximum of $10\%$ of the validated data. 
Next, we constructed the training and validation sets from the rest of the dataset. 
With the data size of many existing datasets being around $20{\sim}30$ 
hours~\cite{DBLP:journals/corr/WangZ15e, THUYG-20}, we downsampled the training data to the extent of 30 hours per language on average to simulate a low-resource setting. As a result, the total size of training data changed from $5,492$ hours to $450$ hours, with the general downsampling ratio $\lambda=0.082$. 
Different downsampling ratios are used for different languages to counter the imbalance of data size among languages.
For a set of languages $\{L_1, ...,L_m\}$ with their proportion $\{p_1, ...,p_m\}$, the downsampling ratio for language $L_i$ is obtained by $ \lambda_i = \frac{p_i^{\alpha-1}}{\sum_{j=1}^{m} p_j^\alpha}\lambda$,  where $\alpha$ is a smooth coefficient that we set to $0.5$.
The size of the validation set is $12.5\%$ of the training set across all languages.
The detailed statistics are shown in Table \ref{tab:data}.

We tokenized the transcriptions at the character level, meaning that the vocabulary is made up solely of individual characters. This also follows \cite{liu2023hierarchical} which verified that character-level tokenization outperforms subword-level (such as BPE) tokenization.

\begin{table}[ht] %\footnotesize %
\setlength{\abovecaptionskip}{10pt}
\begin{center}
{
\caption{Dataset statistics.}
\label{tab:data}
\begin{tabular}{|l|l||r|r|r|}
\hline
% & & \multicolumn{2}{|c|}{Dateset}  \\
% \cline{3-4}
Group & Language & Training & Validation & Test  \\
 &  & (Hours) & (Hours) & (Hours)  \\
%\hhline{|=|=|=|=|}
\hline
\hline
 & Catalan (ca) & 76.3  & 9.5 & 15.3  \\
\cline{2-5}
 & Spanish (es) & 37.6  & 4.7 &  14.4  \\
\cline{2-5}
Romance & French (fr) & 55.7 & 7.0 &  13.4  \\
\cline{2-5}
 & Italian (it) & 33.4  & 4.2 &  15.0  \\
\cline{2-5}
 & Portugal (pt) & 20.7  & 2.6 &  11.4  \\
\hline
 & Belarusian (be) & 63.0  & 7.9 &  13.9  \\
\cline{2-5}
 & Czech (cs) & 42.5  & 5.3 &  5.8  \\
\cline{2-5}
Slavic & Polish (pl) & 37.8 & 4.7 &  12.3  \\
\cline{2-5}
 & Russian (ru) & 27.3  & 3.4 &  14.5  \\
\cline{2-5}
 & Ukrainian (uk) & 23.4  & 2.9 &  7.2  \\
\hline
 & Bashkir (ba) & 29.6  & 3.7 &  12.6  \\
\cline{2-5}
 & Kyrgyz (ky) & 11.3 & 1.4 &  3.8  \\
\cline{2-5}
Turkic & Turkish (tr) & 16.9 & 2.1  &  8.4  \\
\cline{2-5}
 & Tatar (tt) & 10.0  & 1.2 &  3.0  \\
\cline{2-5}
 & Uzbek (uz) & 18.1  & 2.3 &  10.4  \\
\hline
\noalign{\vskip 20pt} 
\end{tabular}
}\\
\end{center}
\vspace{-20pt}
\end{table}

\begin{table*}[t] %\footnotesize %
\setlength{\abovecaptionskip}{10pt}
\begin{center}
{
\caption{CER\% with different clustering methods on the validation set mixing all 15 languages.}
\label{tab:cluster}
\begin{tabular}{|c|c|c||r|r|r|r|r|}
\hline
\multicolumn{2}{|c|}{Clustering Method} & Metric & XLM-base & XLM-large & XLM-XL & LABSE & Mono-Map   \\
\hline
\hline
 &  & Euclidian & 8.7 & 9.4 & 9.0 & 9.1 & 9.4   \\
 \cline{3-8}
 &  & S-Euclidean & 8.7 & 9.2 &8.9 & 9.4 & 9.0   \\
  \cline{3-8}
 & Average & Cityblock & 9.6 & 8.7 &9.2 & 8.5 & \textbf{8.6}   \\
  \cline{3-8}
 &  & Cosine & 8.9 & 8.9 &8.7 & 9.5 & 9.2   \\
  \cline{3-8}
 &  & Correlation & 8.8 & 8.9 &8.8 & 8.5 & 8.7   \\
  \cline{2-8}
  &  & Euclidian & 9.3 & 9.7 &9.2 & 8.7 & 8.7   \\
   \cline{3-8}
Agglomerative &  & S-Euclidean & 9.5 & 8.8 &8.9 & 9.0 & 8.9   \\
  \cline{3-8}
 & Weighted & Cityblock & 8.8 & 9.3 &8.8 & 8.6 & 9.7   \\
  \cline{3-8}
 &  & Cosine & 9.1 & 8.8 &9.0 & 9.1 & 9.4   \\
  \cline{3-8}
 &  & Correlation & 9.1 & 9.2 &\textbf{8.5} & 8.5 & 9.6   \\
  \cline{2-8}
  & Centroid & Euclidian & 8.8 & 9.0 &8.9 & 8.9 & 8.9   \\
    \cline{2-8}
 & Median & Euclidian & 8.9 & 9.0 &9.0 & \textbf{8.5} & 8.7   \\
   \cline{2-8}
 & Ward & Euclidian & 9.4 & 8.6 &8.7 & 8.5 & 8.8   \\
 \hline
 & 2-Means & Euclidian & 10.1 & 9.2 &8.9 & 9.2 & 8.9   \\
   \cline{2-8}
 & Spherical & Cosine & 9.8 & 8.6 &8.7 & 8.9 & 9.3   \\
   \cline{2-8}
  &  & Euclidian & 9.5 & 9.4 &9.7 & 9.2 & 8.7   \\
   \cline{3-8}
Divisive &  & S-Euclidean & \textbf{8.7} & \textbf{8.5} &8.8 & 9.4 & 8.7   \\
  \cline{3-8}
 & 2-medoids & Cityblock & 8.8 & 8.8 &9.2 & 8.8 & 8.7   \\
  \cline{3-8}
 &  & Cosine & 8.9 & 8.9 &9.7 & 8.8 & 8.8   \\
  \cline{3-8}
 &  & Correlation & 9.4 & 9.2 & 9.3 & 9.1 & 8.6   \\
 \hline
 \noalign{\vskip 20pt} 
\end{tabular}
}\\
\end{center}
\vspace{-20pt}
\end{table*}

\iffalse
\begin{table*}[t] %\footnotesize %
%\setlength{\abovecaptionskip}{20pt}
\setlength{\abovecaptionskip}{10pt}
\begin{center}
{
\caption{CER\% by different models trained with languages within the same language group.}
\label{tab:5-lang}
\begin{tabular}{|c|c||r|r|r|}
\hline
 &  &  & \multicolumn{2}{|c|}{H-Softmax}   \\
\cline{4-5}
Training & Test & Softmax &  & \multicolumn{1}{|c|}{Embedding}    \\
\cline{5-5}
 &  &  & Huffman  & XLM-XL \\
  &  &  &  & Wtd Corr \\
\hline
\hline
&ca  & 7.0 & 6.9 & 6.9\\
\cline{2-5}
&es  & 13.1 & 13.0 &12.9 \\
\cline{2-5}
Romance & fr  & 23.3 & 23.3 &  23.0 \\
\cline{2-5}
&it  & 11.4 & 11.5 &  11.6 \\
\cline{2-5}
&pt  & 14.9 & 15.1 &  14.8 \\
\cline{2-5}
&Average  & 13.9 & 14.0 & 13.8 \\
\hline
&be  & 7.6 & 9.9 &7.4 \\
\cline{2-5}
&cs  & 24.1 & 28.3  & 23.0 \\
\cline{2-5}
Slavic & pl  & 17.4 & 19.7 & 16.6 \\
\cline{2-5}
&ru  & 20.6 & 24.1 &  19.3 \\
\cline{2-5}
&uk  & 19.1 & 23.4  & 18.5 \\
\cline{2-5}
&Average  & 17.8 & 21.1  & 17.0 \\
\hline
&ba  & 5.9 & 6.6  & 8.0  \\
\cline{2-5}
&tr  & 29.3 & 31.0  & 30.5  \\
\cline{2-5}
Turkic &tt  &  35.4 & 37.6 & 44.7  \\
\cline{2-5}
&ug  & 17.7 & 19.5 & 22.9  \\
\cline{2-5}
&uz  & 36.9 & 29.6  & 27.3  \\
\cline{2-5}
&Average  & 25.0 & 24.9 &  26.7\\
\hline
tr &tr  & 57.0 & 58.8 &  55.2 \\
\hline
\multicolumn{2}{|c||}{Average} & 11.6 & 10.3 & 10.3  \\
\hline
\noalign{\vskip 20pt} 
\end{tabular}
}\\
\end{center}
\vspace{-20pt}
\end{table*}
\fi

\begin{table*}[!h] %\footnotesize %
\setlength{\abovecaptionskip}{10pt}
\begin{center}
{
\caption{CER\% by different models trained with languages within the same language group. \revised{$\dag$ and $\ddag$ indicate that the results are significantly better than Softmax and Huffman-based H-Softmax at $p<0.05$ based on two-proportion $z$-test, respectively. As test-set sizes vary across languages, we report both ``Global'' and ``Average'' scores. ``Global'' evaluates the concatenated test sets and is used for significance testing, whereas ``Average'' is the unweighted mean across languages, providing a descriptive summary that gives each language equal weight.}}
\label{tab:5-lang}
\begin{tabular}{|c|c||r|r|r|r|r|r|r|}
\hline
 &  &  & \multicolumn{6}{|c|}{H-Softmax}   \\
\cline{4-9}
Training & Test & Softmax &  & \multicolumn{5}{|c|}{Embedding}    \\
\cline{5-9}
 &  &  & Huffman & XLM-base & XLM-large & XLM-XL & LABSE & Mono-Map \\
  &  &  &  & 2-med S-Euc & 2-med S-Euc &Wtd Corr & Median Euc & Avg CB \\
\hline\hline
&ca  & 6.8 & 5.4\rlap{\revised{$^\dag$}} & \textbf{5.1}\rlap{\revised{$^{\dag\ddag}$}} & \textbf{5.1}\rlap{\revised{$^{\dag\ddag}$}} &5.5\rlap{\revised{$^\dag$}}& 5.2\rlap{\revised{$^{\dag\ddag}$}} & 5.5\rlap{\revised{$^\dag$}}\\
\cline{2-9}
&es  & 10.6 & \textbf{8.6}\rlap{\revised{$^\dag$}} & 9.0\rlap{\revised{$^\dag$}} & 9.1\rlap{\revised{$^\dag$}} &8.9\rlap{\revised{$^\dag$}}& 9.9\rlap{\revised{$^\dag$}} &  9.3\rlap{\revised{$^\dag$}}\\
\cline{2-9}
Romance & fr  & 13.0 & 11.3\rlap{\revised{$^\dag$}} & 11.2\rlap{\revised{$^\dag$}} & 11.3\rlap{\revised{$^\dag$}} &12.1\rlap{\revised{$^\dag$}}& \textbf{11.1}\rlap{\revised{$^{\dag\ddag}$}} &  11.6\rlap{\revised{$^\dag$}}\\
\cline{2-9}
&it  & 10.6 & 8.9\rlap{\revised{$^\dag$}} & 9.1\rlap{\revised{$^\dag$}} & 9.3\rlap{\revised{$^\dag$}} & \textbf{8.5}\rlap{\revised{$^{\dag\ddag}$}} & \textbf{8.1}\rlap{\revised{$^{\dag\ddag}$}} &  8.4\rlap{\revised{$^{\dag\ddag}$}}\\
\cline{2-9}
&pt  & 11.1 & 10.3\rlap{\revised{$^\dag$}} & 9.5\rlap{\revised{$^{\dag\ddag}$}} & \textbf{9.2}\rlap{\revised{$^{\dag\ddag}$}} &10.1\rlap{\revised{$^{\dag\ddag}$}}& 9.4\rlap{\revised{$^{\dag\ddag}$}} &  9.5\rlap{\revised{$^{\dag\ddag}$}}\\
\cline{2-9}
&\revised{Global}  & \revised{10.4} & \revised{8.8\rlap{$^\dag$}} & \revised{\textbf{8.7}\rlap{$^{\dag\ddag}$}} & \revised{8.8\rlap{$^\dag$}} & \revised{8.9\rlap{$^\dag$}} & \revised{\textbf{8.7}\rlap{$^{\dag\ddag}$}} & \revised{8.8\rlap{$^\dag$}}\\
\cline{2-9}
&Average  & 10.4 & 8.9 & 8.8 & 8.8 &9.0 & \textbf{8.7} &  8.9\\
\hline
&be  & 5.1 & 4.7\rlap{\revised{$^\dag$}} & \textbf{4.2}\rlap{\revised{$^{\dag\ddag}$}} &4.4\rlap{\revised{$^{\dag\ddag}$}}&4.4\rlap{\revised{$^{\dag\ddag}$}} & 4.3\rlap{\revised{$^{\dag\ddag}$}} & 4.4\rlap{\revised{$^{\dag\ddag}$}}\\
\cline{2-9}
&cs  & 5.9 & 5.7\rlap{\revised{$^\dag$}} & \textbf{5.4}\rlap{\revised{$^{\dag\ddag}$}} & \textbf{5.4}\rlap{\revised{$^{\dag\ddag}$}} &6.0 & 5.5\rlap{\revised{$^{\dag\ddag}$}} & 5.7\rlap{\revised{$^\dag$}}\\
\cline{2-9}
Slavic & pl  & 9.2 & 8.5\rlap{\revised{$^\dag$}} & 8.3\rlap{\revised{$^{\dag\ddag}$}} & \textbf{8.1}\rlap{\revised{$^{\dag\ddag}$}} &9.0\rlap{\revised{$^\dag$}}& 8.6\rlap{\revised{$^\dag$}} & 8.4\rlap{\revised{$^\dag$}}\\
\cline{2-9}
&ru  & 9.8 & \textbf{6.7}\rlap{\revised{$^\dag$}} & \textbf{6.7}\rlap{\revised{$^\dag$}} & \textbf{6.7}\rlap{\revised{$^\dag$}} &7.0\rlap{\revised{$^\dag$}}& \textbf{6.7}\rlap{\revised{$^\dag$}} & 7.4\rlap{\revised{$^\dag$}}\\
\cline{2-9}
&uk  & 12.6 & 10.2\rlap{\revised{$^\dag$}} & \textbf{10.0}\rlap{\revised{$^{\dag\ddag}$}} &10.2\rlap{\revised{$^\dag$}} &11.1\rlap{\revised{$^\dag$}}& 10.5\rlap{\revised{$^\dag$}} & 10.6\rlap{\revised{$^\dag$}}\\
\cline{2-9}
&\revised{Global}  & \revised{8.4} & \revised{6.9\rlap{$^\dag$}} & \revised{\textbf{6.7}\rlap{$^{\dag\ddag}$}} & \revised{\textbf{6.7}\rlap{$^{\dag\ddag}$}} & \revised{7.2\rlap{$^\dag$}} & \revised{6.8\rlap{$^{\dag\ddag}$}} & \revised{7.1\rlap{$^\dag$}}\\
\cline{2-9}
&Average  & 8.5 & 7.2 & \textbf{6.9} &7.0 & 7.5 & 7.1 & 7.3\\
\hline
&ba  & 17.5 & 14.6\rlap{\revised{$^\dag$}} & 14.8\rlap{\revised{$^\dag$}} &16.3\rlap{\revised{$^\dag$}} & 14.1\rlap{\revised{$^{\dag\ddag}$}} & \textbf{13.9}\rlap{\revised{$^{\dag\ddag}$}} & 15.2\rlap{\revised{$^\dag$}} \\
\cline{2-9}
&ky  & 19.8 & 20.8 & 18.6\rlap{\revised{$^{\dag\ddag}$}} &17.7\rlap{\revised{$^{\dag\ddag}$}} & 18.1\rlap{\revised{$^{\dag\ddag}$}} & \textbf{17.3}\rlap{\revised{$^{\dag\ddag}$}} & 18.2\rlap{\revised{$^{\dag\ddag}$}} \\
\cline{2-9}
Turkic &tr  &  13.6 & 12.8\rlap{\revised{$^\dag$}} & 12.6\rlap{\revised{$^{\dag\ddag}$}} &12.7\rlap{\revised{$^\dag$}} & 13.2\rlap{\revised{$^\dag$}} & 13.2\rlap{\revised{$^\dag$}} & \textbf{12.5}\rlap{\revised{$^{\dag\ddag}$}} \\
\cline{2-9}
&tt  & 10.1 & 9.6\rlap{\revised{$^\dag$}} & \textbf{9.0}\rlap{\revised{$^{\dag\ddag}$}} & 9.8\rlap{\revised{$^\dag$}}& 9.1\rlap{\revised{$^{\dag\ddag}$}} & 9.6\rlap{\revised{$^\dag$}} & 9.2\rlap{\revised{$^{\dag\ddag}$}} \\
\cline{2-9}
&uz  & 18.8 & 16.8\rlap{\revised{$^\dag$}} & 17.1\rlap{\revised{$^\dag$}} & \textbf{16.5}\rlap{\revised{$^{\dag\ddag}$}} &17.2\rlap{\revised{$^\dag$}}& 17.2\rlap{\revised{$^\dag$}} & 16.9\rlap{\revised{$^\dag$}} \\
\cline{2-9}
&\revised{Global}  & \revised{16.6} & \revised{14.9\rlap{$^\dag$}} & \revised{14.8\rlap{$^\dag$}} & \revised{15.1\rlap{$^\dag$}} & \revised{14.7\rlap{$^{\dag\ddag}$}} & \revised{\textbf{14.6}\rlap{$^{\dag\ddag}$}} & \revised{14.8\rlap{$^\dag$}} \\
\cline{2-9}
&Average  & 16.0 & 14.9 & 14.4 & 14.6& 14.3 & \textbf{14.2} & 14.4 \\
\hline
\multicolumn{2}{|c||}{\revised{Global}} & \revised{11.1} & \revised{9.5\rlap{$^\dag$}} & \revised{\textbf{9.3}\rlap{$^{\dag\ddag}$}} & \revised{9.4\rlap{$^\dag$}} & \revised{9.6\rlap{$^\dag$}} & \revised{\textbf{9.3}\rlap{$^{\dag\ddag}$}} & \revised{9.5\rlap{$^\dag$}} \\
\hline
\multicolumn{2}{|c||}{Average} & 11.6 & 10.3 & \textbf{10.0} & 10.1 & 10.3 & \textbf{10.0} & 10.2 \\
\hline
\noalign{\vskip 20pt}
\end{tabular}
}\\
\end{center}
\vspace{-20pt}
\end{table*}

\begin{table*}[t] %\footnotesize %
\setlength{\abovecaptionskip}{10pt}
\begin{center}
{
\caption{CER\% by models trained with all the languages across different language groups. \revised{$\dag$ and $\ddag$ indicate that the results are significantly better than Softmax and Huffman-based H-Softmax at $p<0.05$ based on two-proportion $z$-test, respectively. As test-set sizes vary across languages, we report both ``Global'' and ``Average'' scores. ``Global'' evaluates the concatenated test sets and is used for significance testing, whereas ``Average'' is the unweighted mean across languages, providing a descriptive summary that gives each language equal weight.}}
\label{tab:15-lang}
\begin{tabular}{|c|c||r|r|r|r|r|r|r|}
\hline
 &  &  & \multicolumn{6}{|c|}{H-Softmax}   \\
\cline{4-9}
Training & Test & Softmax &  & \multicolumn{5}{|c|}{Embedding}    \\
\cline{5-9}
 &  &  & Huffman & XLM-base & XLM-large & XLM-XL & LABSE & Mono-Map \\
  &  &  &  & 2-med S-Euc & 2-med S-Euc &Wtd Corr & Median Euc & Avg CB \\
\hline\hline
&ca  & 6.9 & 6.0\rlap{\revised{$^\dag$}} & 5.8\rlap{\revised{$^{\dag\ddag}$}} & 5.7\rlap{\revised{$^{\dag\ddag}$}} & 5.7\rlap{\revised{$^{\dag\ddag}$}} & 6.1\rlap{\revised{$^\dag$}} & \textbf{5.6}\rlap{\revised{$^{\dag\ddag}$}}\\
\cline{2-9}
&es  & 10.7 & 9.5\rlap{\revised{$^\dag$}} & 8.3\rlap{\revised{$^{\dag\ddag}$}} & 8.7\rlap{\revised{$^{\dag\ddag}$}} & \textbf{8.2}\rlap{\revised{$^{\dag\ddag}$}} & 9.0\rlap{\revised{$^{\dag\ddag}$}} & \textbf{8.2}\rlap{\revised{$^{\dag\ddag}$}}\\
\cline{2-9}
 & fr  & 13.7 & \textbf{11.7}\rlap{\revised{$^\dag$}} & 11.8\rlap{\revised{$^\dag$}} & \textbf{11.7}\rlap{\revised{$^\dag$}} & 11.9\rlap{\revised{$^\dag$}} & 12.5\rlap{\revised{$^\dag$}} & \textbf{11.7}\rlap{\revised{$^\dag$}}\\
\cline{2-9}
&it  & 8.6 & 8.0\rlap{\revised{$^\dag$}} & 7.7\rlap{\revised{$^{\dag\ddag}$}} & \textbf{7.4}\rlap{\revised{$^{\dag\ddag}$}} & 7.6\rlap{\revised{$^{\dag\ddag}$}} & \textbf{7.4}\rlap{\revised{$^{\dag\ddag}$}} &  8.1\rlap{\revised{$^\dag$}}\\
\cline{2-9}
&pt  & 37.6 & 29.0\rlap{\revised{$^\dag$}} & 11.5\rlap{\revised{$^{\dag\ddag}$}} & 12.1\rlap{\revised{$^{\dag\ddag}$}} & 12.7\rlap{\revised{$^{\dag\ddag}$}} & \textbf{10.7}\rlap{\revised{$^{\dag\ddag}$}} &  12.1\rlap{\revised{$^{\dag\ddag}$}}\\
\cline{2-9}
&\revised{Romance Global}  & \revised{13.9} & \revised{11.7\rlap{$^\dag$}} & \revised{\textbf{8.8}\rlap{$^{\dag\ddag}$}} & \revised{8.9\rlap{$^{\dag\ddag}$}} & \revised{8.9\rlap{$^{\dag\ddag}$}} & \revised{9.0\rlap{$^{\dag\ddag}$}} & \revised{8.9\rlap{$^{\dag\ddag}$}}\\
\cline{2-9}
&Romance Average  & 15.5 & 12.8 & \textbf{9.0} & 9.1 & 9.2 & 9.1 & 9.1\\
\cline{2-9}
&be  & 5.7 & 6.0 & 6.0 & 5.2\rlap{\revised{$^{\dag\ddag}$}} & 5.0\rlap{\revised{$^{\dag\ddag}$}} & \textbf{4.5}\rlap{\revised{$^{\dag\ddag}$}} & 5.3\rlap{\revised{$^{\dag\ddag}$}}\\
\cline{2-9}
&cs  & 6.0 & 5.7\rlap{\revised{$^\dag$}} & 5.7\rlap{\revised{$^\dag$}} & 6.0 & 6.5 & 6.5 & 6.7\\
\cline{2-9}
All & pl  & 9.0 & 8.5\rlap{\revised{$^\dag$}} & 8.1\rlap{\revised{$^{\dag\ddag}$}} & 8.3\rlap{\revised{$^{\dag\ddag}$}} & \textbf{8.0}\rlap{\revised{$^{\dag\ddag}$}} & 8.8\rlap{\revised{$^\dag$}} & 9.1\\
\cline{2-9}
&ru  & 9.8 & 8.9\rlap{\revised{$^\dag$}} & \textbf{8.3}\rlap{\revised{$^{\dag\ddag}$}} & 8.5\rlap{\revised{$^{\dag\ddag}$}} & 8.4\rlap{\revised{$^{\dag\ddag}$}} & 10.3 & 8.6\rlap{\revised{$^{\dag\ddag}$}}\\
\cline{2-9}
&uk  & 11.8 & 10.3\rlap{\revised{$^\dag$}} & 10.0\rlap{\revised{$^{\dag\ddag}$}} & \textbf{9.7}\rlap{\revised{$^{\dag\ddag}$}} & 9.9\rlap{\revised{$^{\dag\ddag}$}} & 10.2\rlap{\revised{$^\dag$}} & 10.0\rlap{\revised{$^{\dag\ddag}$}}\\
\cline{2-9}
&\revised{Slavic Global}  & \revised{8.4} & \revised{7.9\rlap{$^\dag$}} & \revised{7.6\rlap{$^{\dag\ddag}$}} & \revised{7.5\rlap{$^{\dag\ddag}$}} & \revised{\textbf{7.4}\rlap{$^{\dag\ddag}$}} & \revised{8.1\rlap{$^\dag$}} & \revised{7.8\rlap{$^{\dag\ddag}$}}\\
\cline{2-9}
&Slavic Average  & 8.5 & 7.9 & 7.6 & \textbf{7.5} & 7.6 & 8.1 & 7.9\\
\cline{2-9}
&ba  & 12.4 & 10.8\rlap{\revised{$^\dag$}} & 13.7 & \textbf{10.6}\rlap{\revised{$^{\dag\ddag}$}} & 11.1 & 10.9\rlap{\revised{$^\dag$}} & 12.1\rlap{\revised{$^\dag$}} \\
\cline{2-9}
&ky  & 15.2 & \textbf{13.9}\rlap{\revised{$^\dag$}} & 16.0 & 16.3 & 15.8 & \textbf{13.9}\rlap{\revised{$^\dag$}} & 16.6 \\
\cline{2-9}
 &tr  & 13.7 & 11.0\rlap{\revised{$^\dag$}} & 10.9\rlap{\revised{$^\dag$}} & 11.2\rlap{\revised{$^\dag$}} & 11.7\rlap{\revised{$^\dag$}} & \textbf{10.8}\rlap{\revised{$^{\dag\ddag}$}} & \textbf{10.8}\rlap{\revised{$^{\dag\ddag}$}} \\
\cline{2-9}
&tt  & 6.7 & 7.7 & 6.9\rlap{\revised{$^\ddag$}} & 7.0\rlap{\revised{$^\ddag$}} & 6.5\rlap{\revised{$^\ddag$}} & 6.9\rlap{\revised{$^\ddag$}} & 6.7\rlap{\revised{$^\ddag$}} \\
\cline{2-9}
&uz  & 14.7 & 13.5\rlap{\revised{$^\dag$}} & 13.8\rlap{\revised{$^\dag$}} & 13.5\rlap{\revised{$^\dag$}} & 13.3\rlap{\revised{$^{\dag\ddag}$}} & \textbf{12.6}\rlap{\revised{$^{\dag\ddag}$}} & \textbf{12.6}\rlap{\revised{$^{\dag\ddag}$}} \\
\cline{2-9}
&\revised{Turkic Global}  & \revised{13.1} & \revised{11.6\rlap{$^\dag$}} & \revised{12.7\rlap{$^\dag$}} & \revised{11.7\rlap{$^\dag$}} & \revised{11.9\rlap{$^\dag$}} & \revised{\textbf{11.3}\rlap{$^{\dag\ddag}$}} & \revised{11.8\rlap{$^\dag$}}\\
\cline{2-9}
&Turkic Average  & 12.5 & 11.4 & 12.3 & 11.7 & 11.7 & \textbf{11.0} & 11.8 \\
\hline
\multicolumn{2}{|c||}{\revised{Global}} & \revised{11.8} & \revised{10.4\rlap{$^\dag$}} & \revised{9.2\rlap{$^{\dag\ddag}$}} & \revised{\textbf{9.0}\rlap{$^{\dag\ddag}$}} & \revised{9.1\rlap{$^{\dag\ddag}$}} & \revised{9.2\rlap{$^{\dag\ddag}$}} & \revised{9.2\rlap{$^{\dag\ddag}$}} \\
\hline
\multicolumn{2}{|c||}{Average} & 12.2 & 10.7 & 9.6 & 9.5 & 9.5 & \textbf{9.4} & 9.6 \\
\hline
\noalign{\vskip 20pt} 
\end{tabular}
}\\
\end{center}
\vspace{-20pt}
\end{table*}

\begin{table}[t] %\footnotesize %
\setlength{\abovecaptionskip}{10pt}
\begin{center}
{
\caption{\revised{Real Time Factor (RTF) by different models with the same computational resource. The training and test sets are 5 Romance languages. }}
\label{tab:rtf}
\begin{tabular}{|c||r|r|r|}
\hline
 &  &  \multicolumn{2}{|c|}{\revised{H-Softmax}}   \\
\cline{3-4}
  & \revised{Softmax} &  & \multicolumn{1}{|c|}{\revised{Embedding}}    \\
\cline{4-4}
 &  & \revised{Huffman} &  \revised{LABSE}  \\
  &  &  & \revised{Median Euc}  \\
\hline\hline
  \revised{RTF} & \revised{\textbf{0.156}} & \revised{0.164}  & \revised{0.240} \\
   \cline{1-4}
   \revised{Tree Depth} & \revised{-} & \revised{26} & \revised{64} \\
\cline{1-4}
\hline
\noalign{\vskip 20pt}
\end{tabular}
}\\
\end{center}
\vspace{-20pt}
\end{table}

\begin{table}[t] %\footnotesize %
\setlength{\abovecaptionskip}{10pt}
\begin{center}
{
\caption{\revised{CER\% by different models trained with monolingual data. $\dag$ and $\ddag$ indicate that the results are significantly better than Softmax and Huffman-based H-Softmax at $p<0.05$ based on two-proportion $z$-test, respectively. As test-set sizes vary across languages, we report both ``Global'' and ``Average'' scores. ``Global'' evaluates the concatenated test sets and is used for significance testing, whereas ``Average'' is the unweighted mean across languages, providing a descriptive summary that gives each language equal weight.}}
\label{tab:mono}
\begin{tabular}{|c|c||r|r|r|}
\hline
 &  &  & \multicolumn{2}{|c|}{\revised{H-Softmax}}   \\
\cline{4-5}
\revised{Training} & \revised{Test} & \revised{Softmax} &  & \multicolumn{1}{|c|}{\revised{Embedding}}    \\
\cline{5-5}
 &  &  & \revised{Huffman} & \revised{LABSE} \\
  &  &  &  & \revised{Median Euc} \\
\hline\hline
\revised{ca} & \revised{ca}  & \revised{12.4} & \revised{8.1\rlap{$^\dag$}} & \revised{\textbf{8.0}\rlap{$^\dag$}}\\
\hline
\revised{es} & \revised{es}  & \revised{16.1} & \revised{\textbf{14.5}\rlap{$^\dag$}} &  \revised{15.4\rlap{$^\dag$}} \\
\hline
\revised{fr} & \revised{fr}  & \revised{19.0} & \revised{17.3\rlap{$^\dag$}} & \revised{\textbf{16.8}\rlap{$^{\dag\ddag}$}} \\
\hline
\revised{it} & \revised{it}  & \revised{16.4} & \revised{13.5\rlap{$^\dag$}} & \revised{\textbf{13.2}\rlap{$^{\dag\ddag}$}}\\
\hline
\revised{pt} & \revised{pt}  & \revised{28.1} & \revised{26.7\rlap{$^\dag$}} & \revised{\textbf{26.3}\rlap{$^{\dag\ddag}$}}  \\
\hline
\multicolumn{2}{|c||}{\revised{Global}}  & \revised{17.7} & \revised{\textbf{15.2}\rlap{$^\dag$}} & \revised{\textbf{15.2}\rlap{$^\dag$}} \\
\hline
\multicolumn{2}{|c||}{\revised{Average}}  & \revised{18.4} & \revised{16.0} & \revised{\textbf{15.9}}\\
\hline
\noalign{\vskip 20pt}
\end{tabular}
}\\
\end{center}
\vspace{-20pt}
\end{table}

\subsection{Model Training Details} 

%NOTE: Mainly copied from ICASSP paper

For acoustic features, the 80-dimensional log-Mel filterbanks (FBANK) are computed 
with a $25$ms window and a 10ms shift. Besides, SpecAugment \cite{DBLP:conf/interspeech/ParkCZCZCL19} is applied to $2$ frequency masks with maximum frequency mask $F = 10$ and 2-time masks with maximum time mask $T = 50$ to alleviate over-fitting. 

Both H-Softmax and Softmax models are trained using the same base network structure with a 12-layer conformer~\cite{DBLP:conf/interspeech/GulatiQCPZYHWZW20} encoder and a 6-layer transformer~\cite{vaswani2017attention} decoder. 
Two convolution sub-sampling layers with kernel size $3\times3$ and stride $2$ are used in the front of the encoder.  
The networks are constructed using WeNet toolkit \cite{yao21_interspeech}.
%\footnote{Our implementation is  https://github.com/Derek-Gong/hSoftmax\label{github}} 

Adam optimizer is used with a learning rate schedule with $25,000$ warm-up steps. 
The initial learning rate is $0.00005$. 
The maximum number of epochs is $100$.
We kept checkpoints after each epoch and used the average of $5$ with the best validation accuracy for testing.

\subsection{Preliminary Comparison of Different Clustering Configuration}

%TODO: One language group with two settings
%also with different distance metrics

To determine the optimal clustering method and distance metric for our study, we conducted a series of preliminary experiments. These experiments were designed to evaluate the effectiveness of various clustering configurations with different sources of cross-lingual embeddings, recognizing that each embedding might work best with a specific clustering configuration. We conducted training on the whole training set and evaluation on the whole validation set across all 15 languages in our dataset. Based on the results in Table \ref{tab:cluster}, we selected the most effective clustering configuration for each embedding source for our main experiment: 2-medoids clustering with standardized Euclidean distance for XLM-base and XLM-large, weighted clustering with correlation distance for XLM-XL, median clustering for LABSE, and average clustering with city block distance for Mono-Map.

\subsection{Main Results}

Our experiments were structured around two distinct scenarios. The first one focuses on five languages all belonging to a single language group. This setup aims to explore the effectiveness of embedding-based H-Softmax within a group of closely related languages, where phonetic and syntactic similarities might influence the performance of ASR systems. 
%The results of this scenario are shown in Table \ref{tab:5-lang}. 
The second scenario broadens the scope to include 15 languages spanning multiple language groups. This more complex arrangement tests the robustness and adaptability of our methods across a broader linguistic spectrum, particularly examining how well the proposed method handles linguistic diversity compared to previous methods. 
%The results are presented in Table \ref{tab:15-lang}.

%Our experiments were structured around two distinct scenarios: one involving five languages from a single language family, whose results are shown in Table \ref{tab:5-lang}, and another encompassing 15 languages from diverse language families, whose results are shown in Table \ref{tab:15-lang}. 

\revised{The results for the two scenarios are reported in Tables \ref{tab:5-lang} and \ref{tab:15-lang}. In the first scenario, two embedding-based methods (XLM-base and LaBSE) significantly outperform the Huffman-based method on global CER. In the second scenario, all the embedding-based methods significantly outperform the Huffman baseline on global CER. When considering the average CERs, nearly all embedding-based methods outperform the Huffman-based method in both scenarios (the only exception is XLM-XL in the first scenario). These results indicate a clear advantage of the embedding-based method over the Huffman baseline. The larger gains observed in the second scenario might be due to the fact that the best clustering methods were determined based on the performance across all 15 languages, which is the same as the second scenario.}

\revised{When comparing embedding sources, LABSE, XLM-large, and XLM-XL demonstrated superior performance over XLM-base in the second scenario (a pattern not seen in the first), indicating the benefits of utilizing larger embedding models in multilingual ASR tasks involving a larger number of languages.
Pre-trained embeddings generally outperform Mono-Map in both scenarios, though the margin is relatively narrow. This confirms the value of sophisticated pre-trained models in enhancing multilingual ASR performance. Moreover, the H-Softmax method significantly surpasses the traditional Softmax method in overall performance, confirming its superiority across both embedding-based and Huffman-based approaches.}

\subsection{\revised{Inference Speed}}

\revised{
We report inference speed in Table \ref{tab:rtf}. Compared with Softmax and the Huffman-based H-Softmax, the embedding-based variants produce higher inference latency. This slowdown arises from the deeper trees produced by embedding clustering: unlike Huffman coding, which explicitly minimizes average path length, the embedding-based construction does not optimize for shallow paths. This is a limitation of the embedding-based method, making it better suited to offline use or deployments on high-performance hardware.}

\subsection{\revised{Monolingual Training}}

\revised{
Although our primary motivation is multilingual knowledge transfer, the proposed method is also applicable to monolingual data. We therefore conducted monolingual experiments as an ablation. As shown in Table \ref{tab:mono}, the resulting CERs are notably worse than those in Tables \ref{tab:5-lang} and \ref{tab:15-lang}, showing the effective cross-lingual transfer in our multilingual setups.}

\revised{Within the monolingual setting, Huffman-based H-Softmax still significantly outperforms the Softmax baseline; however, the additional gain of the embedding-based H-Softmax (with LaBSE) over Huffman becomes marginal. This indicates that while the H-Softmax architecture is beneficial even monolingually, the extra improvements from semantically clustered trees depend primarily on cross-lingual sharing.}

%\vspace{57pt}

%These experiments demonstrate that the H-Softmax approach significantly surpasses the traditional Softmax method in performance. Moreover, we observed that clustering based on embeddings marginally outperforms the Huffman-based approach, demonstrating the effectiveness of our proposed method. 

%When comparing embedding sources, pre-trained embeddings showed a slight advantage over Mono-Map. In the scenario with five languages, there was a negligible difference in performance between XLM and LABSE embeddings. However, in the more extensive 15-language setup, LABSE, XLM-large, and XLM-XL demonstrated superior performance over XLM-base, indicating the benefits of utilizing larger embedding models in multilingual ASR tasks involving a larger number of languages.
% \begin{figure*}[ht]
%     \centering
%     \begin{minipage}[b]{0.32\textwidth}
%         \centering
%         \includegraphics[width=\textwidth]{labse_correlation_plot.pdf}
%     \end{minipage}
%     \hfill
%     \begin{minipage}[b]{0.32\textwidth}
%         \centering
%         \includegraphics[width=\textwidth]{mono-map_correlation_plot.pdf}
%     \end{minipage}
%     \hfill
%     \begin{minipage}[b]{0.32\textwidth}
%         \centering
%         \includegraphics[width=\textwidth]{xlm-base_correlation_plot.pdf}
%     \end{minipage}
%     \caption{Correlation plots of CER\% and Bilingual Language Induction p@1 on Romance Languages for LABSE, Mono-Map, and XLM-base}
%     \label{fig:correlation_plots}
% \end{figure*}
% \section{Analysis}

\begin{figure*}[htb]
  \setlength{\abovecaptionskip}{15pt}
  \centering
  \includegraphics[width=0.9\textwidth]{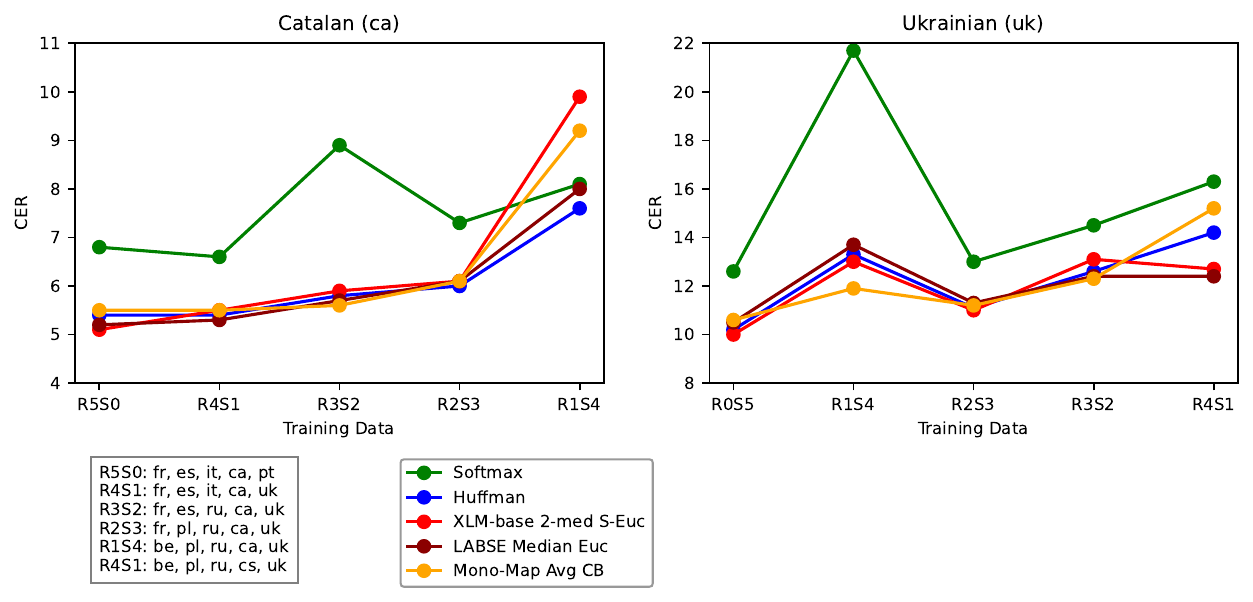}
  \caption{CER\% on Catalan and Ukrainian with different proportions of languages from the same group in the training data. The composition of each training data is shown in the box at the bottom-left corner. }
  \label{fig:cauk}
\end{figure*}

\begin{figure*}[ht]
    \centering
\includegraphics[width=0.9\textwidth]{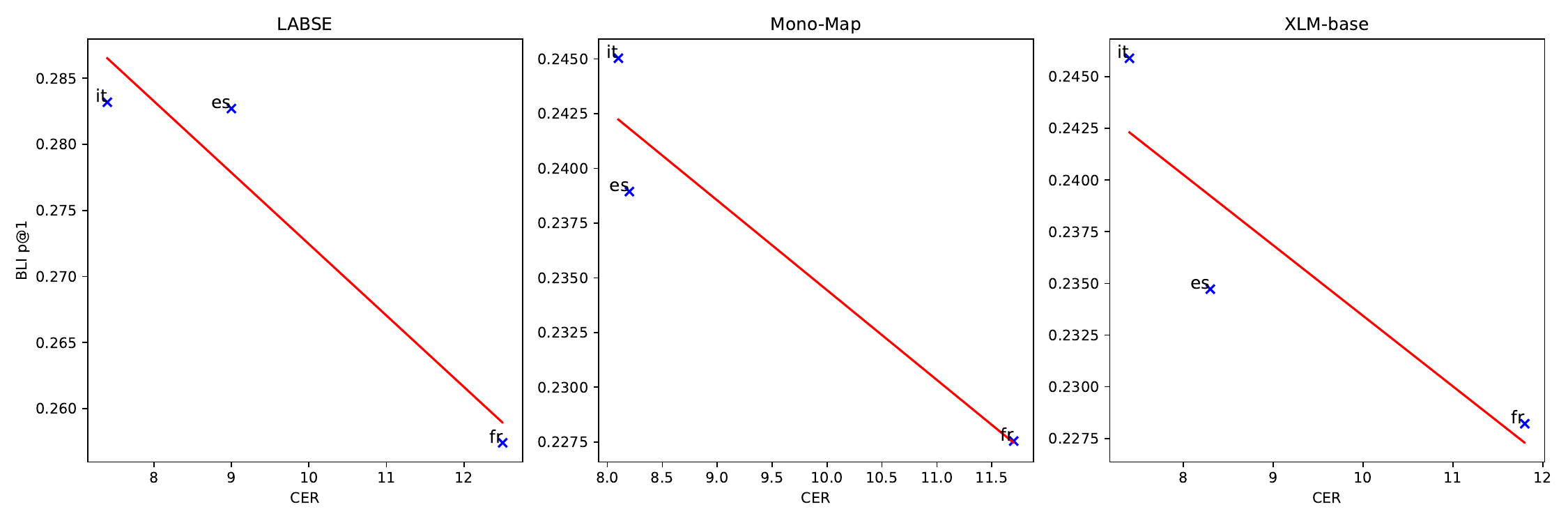}
    \caption{Correlation plots of CER\% and Bilingual Language Induction p@1 on Romance Languages for LABSE, Mono-Map, and XLM-base. The X-axis represents the ASR performance of the language. The Y-axis represents the BLI p@1 score of each language as source language. The blue dots represents the performance of each language. The red line in the figure represents the linear regression fit applied to the data points across languages.}
    \label{fig:correlation_plots}
\end{figure*}

\section{Analysis}

%\subsection{tree distance of the same token in different languages}

%TODO: Calculate it for Slavic and Turkic
\subsection{Tree Structure of Different Construction Methods}

To validate whether the embedding-based method generates a tree structure that better captures the similarity between different tokens compared to the Huffman-based method, we examined the structures produced by these different construction methods. As a result, intuitively, similar tokens are indeed positioned closer to each other in the tree generated by the embedding-based method, especially in the LABSE-based tree. Table \ref{tab:tree} in the Appendix shows an example comparing the Huffman-based and LABSE-based trees in experiments where models were trained with all 15 languages. Due to space limitations, we only selected a portion of the vocabulary, specifically Latin characters (including accented and special characters). The positions of tokens in the tree are represented using Huffman-code-like representations, with the vocabulary ordered based on a depth-first traversal of the tree. In the LABSE-based tree, it is evident that similar tokens (such as ``O" and its accented versions) cluster together, which does not happen in the Huffman-based tree. This confirms that hierarchical clustering indeed constructs the desired tree structure.

\subsection{Proportion of Languages from the Same Group}

Although our primary results include experiments trained with languages from the same group as well as languages from different groups, it is difficult to recognize the influence of language relatedness in the training data from the results. This is because the number of languages and the data size vary significantly between these two experimental setups. To further analyze the impact of language relatedness on model performance, we conducted additional experiments with a constant number of languages. We systematically replaced languages from the Romance group with Slavic languages one at a time, thereby creating various combinations of Romance and Slavic languages, as depicted in Figure \ref{fig:cauk}. This substitution process was carefully managed by replacing languages with the same data size ranking within their respective families, ensuring only minimal changes in the overall data size.

We present the results for two specific languages (Catalan and Ukrainian) that feature in most combinations shown in Figure~\ref{fig:cauk}. Generally, the CER tends to increase as more languages from the opposing group are introduced into the training data, which aligns intuitively with expectations. 
For Catalan, the performance degradation using H-Softmax methods is more significant than that of Softmax after the other 4 languages are replaced. This result shows the significance of language relatedness in the effectiveness of H-Softmax, as it relies on capturing cross-lingual knowledge. 
Conversely, this trend is less marked for Ukrainian, where we also note a significant performance drop when just one specific Romance language is introduced. These observations suggest complexities that need further investigation, which we leave as future work.

\subsection{Bilingual Lexicon Induction}

We observe that in the Romance language family, while French has a larger training data size, the ASR performance of the French language is lower than other Romance languages across different settings. To investigate the factors that cause ASR accuracy to vary across different languages, we conduct an analysis of each of the cross-linguistic mapping abilities of the corresponding embedding, specifically through the task of Bilingual Language Induction (BLI). BLI is tested by aligning word embeddings from a source language to their corresponding translations in a target language, and it is evaluated using the precision at rank 1 (p@1), which measures the percentage of correct matches where the highest-ranked target word corresponds to the correct translation of the source word. In our experiments, we used the MUSE~\cite{DBLP:conf/iclr/LampleCRDJ18, DBLP:conf/iclr/LampleCDR18} dataset and focused on the mapping between Romance languages (Spanish, French, and Italian) in both BLI directions, resulting in six experiment settings. The word embeddings used for BLI are formed by averaging character embeddings, which aligns with the hierarchical softmax tree structure in our multilingual ASR model. We computed the mean performance for each individual language when used as the source language. As shown in Figure \ref{fig:correlation_plots}, we present the correlation between the Character Error Rate (CER\%) and the BLI p@1 scores for each embedding. The results indicate that embeddings with higher mapping ability at the character level tend to achieve lower CER\% scores in a multilingual ASR setting. Thus, the well-learned embeddings proposed in our method capture cross-lingual similarities better in Multilingual ASR, resulting in higher performance compared to previous approaches. 
%TODO
%character level induction data has not been formed yet. 
%other settings all set. Test with roman es, fr, pt, it with two directions of 12 experiments. report the mean and individual results. embeddings should be form of average of character embedding 

\begin{table}[t] %\footnotesize %
\setlength{\abovecaptionskip}{10pt}
\begin{center}
{
\caption{\revised{CER\% on Portuguese by models trained without specific languages.}}
\label{tab:remove}
\begin{tabular}{|c|c||r|r|}
\hline
Training & Test & Softmax & H-Softmax (Huffman)   \\
\hline
\hline
All &  & 37.6 & 29.0 \\
\cline{1-1}
\cline{3-4}
All w/o be &  & 12.0 & 11.1\\
\cline{1-1}
\cline{3-4}
All w/o cs &  & 13.2 & 19.9\\
\cline{1-1}
\cline{3-4}
All w/o pl &  & 10.9 & 9.3\\
\cline{1-1}
\cline{3-4}
All w/o ru &  & 12.7 & 12.1\\
\cline{1-1}
\cline{3-4}
All w/o uk & pt & 12.1 & 11.3\\
\cline{1-1}
\cline{3-4}
All w/o ba &  & 10.4 & 11.6\\
\cline{1-1}
\cline{3-4}
All w/o ky &  & 15.6 & 12.4\\
\cline{1-1}
\cline{3-4}
All w/o tr &  & 13.9 & 18.0\\
\cline{1-1}
\cline{3-4}
All w/o tt &  & 15.6 & 15.1\\
\cline{1-1}
\cline{3-4}
All w/o uz &  & 11.0 & 11.2\\

\hline
\noalign{\vskip 20pt} 
\end{tabular}
}\\
\end{center}

\vspace{-20pt}
\end{table}

\subsection{Incorrect Language Prediction}
%\subsection{Error Analysis of Wrong Language and Language ID}

%TODO: mixture of 15 langauges, higher error rate of language identification (check ICASSP pt -10 points)

Table \ref{tab:15-lang} shows that Softmax and Huffman-based H-Softmax perform poorly on Portuguese when training with 15 languages. Upon examining the predicted transcriptions for Portuguese, we discovered that a significant portion (around $10.4\%$ for Softmax and $2.7\%$ for Huffman-based H-Softmax) of the transcriptions were erroneously predicted in the Cyrillic alphabet, whereas Portuguese uses the Latin alphabet. This suggests that the model incorrectly processes many Portuguese utterances as if they were in another language. Note that our multilingual ASR experiments are conducted in a language-agnostic setting, which can occasionally lead to such errors.

To determine whether including a specific language in the training set contributes to this performance degradation for Portuguese, we conduct experiments to remove one of the Slavic or Turkic languages from the training data and observe the outcomes, as detailed in Table \ref{tab:remove}. The performance generally recovers with each language removal, only some of them resulted in less recovery.
%but we noted that removing two languages with smaller data sets (...) resulted in lesser recovery. 
This observation implies that no single language specifically causes the performance drop; rather, the collective data from 15 languages may surpass the model's implicit capacity for language identification.

Conversely, the embedding-based H-Softmax does not exhibit similar performance degradation, suggesting that an embedding-based vocabulary tree can enhance the model's ability to identify languages. By structuring pronunciation identification and language identification in separate layers within the tree, the embedding-based method simplifies the language identification task. As a result, the embedding-based H-Softmax can be more effective than the Huffman-based method in language-agnostic multilingual ASR.

%Tested language-aware results (Table for gold LID?)
%Also with language identification: Table \ref{tab:lid-15}
%It shows that ...

%3 language groups and a mixture

%\subsection{data size and performance, i.e. how low resource would our method be useful}

%TODO: design the experiments 

\section{Conclusion}

In this study, we advanced the decoding stage of low-resource multilingual ASR by proposing a novel embedding-based H-Softmax method. The proposed method leverages hierarchical clustering of cross-lingual embeddings, overcoming the limitations of the earlier Huffman-based method that relies on shallow frequency features. 
Our experimental analysis, spanning languages within and across different families, demonstrated the improvement of our embedding-based H-Softmax over both the traditional Softmax and the Huffman-based method, which confirmed that cross-lingual embeddings can convey more nuanced information than mere frequency.
\revised{In the future, we plan to generalize the proposed method to larger state-of-the-art ASR models, and explore the potential of applying our method to the realm of speech translation.}

\section*{Acknowledgments}
This work was supported by Kyoto University Division of Graduate Studies SPRING Program, and JSPS KAKENHI Grant Number JP23K28144 and JP23K11227.
%This should be a simple paragraph before the References to thank those individuals and institutions who have supported your work on this article.

\bibliographystyle{IEEEtran}
\bibliography{refs}

\newpage

{\appendix[]
\begin{table*}[t] %\footnotesize %
\setlength{\abovecaptionskip}{10pt}
\begin{center}
{
\caption{Examples of Huffman-code-like representations of two vocabulary tree structures
the Huffman-based tree and one of the embedding-based trees from the experiments where models were trained with all 15 languages. Only Latin characters (including accented and special characters) are shown.}
\label{tab:tree}
\scriptsize
\begin{tabular}{|l|l||l|l|}
\hline
\multicolumn{2}{|c||}{Huffman} & \multicolumn{2}{c|}{LABSE Median Euc} \\
\hline
Char & Code & Char & Code   \\
\hline
U & 00000 & X & 1110 \\
D & 00001 & Q & 11110 \\
ú & 00010000111 & ý & 1111111111111110 \\
W & 00010001 & ÿ & 11111111111111110 \\
î & 00010010111011 & Ø & 1111111111111111100 \\
ö & 000100101111 & M & 1111111111111111111111101110101111111111001110 \\
ò & 00010011010 & ò & 1111111111111111111111101110101111111111011111110110 \\
X & 000110110 & ö & 1111111111111111111111101110101111111111011111110111100 \\
È & 00011011111001 & ø & 1111111111111111111111101110101111111111011111110111101 \\
Á & 0001111010000100 & H & 11111111111111111111111011101011111111110111111101111100100 \\
K & 0010010 & R & 11111111111111111111111011101011111111110111111101111101011110 \\
ó & 001001101 & L & 111111111111111111111110111010111111111101111111011111011010100 \\
Y & 0010110 & T & 11111111111111111111111011101011111111110111111101111101110110 \\
H & 0010111 & ñ & 11111111111111111111111011101011111111110111111101111101111010 \\
Q & 00110010 & â & 111111111111111111111110111010111111111101111111011111011110110 \\
õ & 001100110100100 & ã & 1111111111111111111111101110101111111111011111110111110111101110 \\
Î & 00110011010010101000 & ï & 11111111111111111111111011101011111111110111111101111101111011110 \\
æ & 001100110100101010010 & ì & 111111111111111111111110111010111111111101111111011111011110111110 \\
Û & 001100110100101010101111 & õ & 11111111111111111111111011101011111111110111111101111101111011111110 \\
Ü & 00110011100001011 & í & 11111111111111111111111011101011111111110111111101111101111011111111101110010 \\
ÿ & 00110011100010100000101 & à & 1111111111111111111111101110101111111111011111110111110111101111111110111010 \\
Ñ & 001100111000101000100010 & á & 1111111111111111111111101110101111111111011111110111110111101111111110111011 \\
Æ & 0011001110001010001001 & ê & 1111111111111111111111101110101111111111011111110111110111101111111110111100 \\
Ô & 0011001110001010001111 & î & 11111111111111111111111011101011111111110111111101111101111011111111101111011111110111010 \\
Â & 0011001110001010011 & ë & 1111111111111111111111101110101111111111011111110111110111101111111110111101111111011101110 \\
Ö & 0011001110001011 & ä & 111111111111111111111110111010111111111101111111011111011110111111111011110111111101110111100 \\
ñ & 0011001110101 & æ & 111111111111111111111110111010111111111101111111011111011110111111111011110111111101110111101 \\
P & 001110 & å & 11111111111111111111111011101011111111110111111101111101111011111111101111011111110111100 \\
Z & 0011111 & ü & 111111111111111111111110111010111111111101111111011111011110111111111011110111111101111110 \\
A & 0100 & ú & 11111111111111111111111011101011111111110111111101111101111011111111101111011111110111111110 \\
L & 01010 & ù & 111111111111111111111110111010111111111101111111011111011110111111111011110111111101111111110 \\
G & 0101101 & ç & 1111111111111111111111101110101111111111011111110111110111101111111110111101111111111011110 \\
è & 0101110001 & A & 111111111111111111111110111010111111111101111111011111011110111111111110100 \\
ê & 010111010001 & Å & 1111111111111111111111101110101111111111011111110111110111101111111111101110 \\
ü & 01011101011 & Á & 11111111111111111111111011101011111111110111111101111101111011111111111011110 \\
á & 010111011 & À & 111111111111111111111110111010111111111101111111011111011110111111111110111110 \\
E & 0110 & D & 111111111111111111111110111010111111111101111111011111011111110110 \\
é & 01111101 & C & 1111111111111111111111101110101111111111011111110111110111111111000 \\
T & 10000 & Ç & 111111111111111111111110111010111111111101111111011111011111111101110 \\
ù & 10001011000001 & K & 11111111111111111111111011101011111111110111111101111101111111111000 \\
ë & 10001011001100000 & Ê & 11111111111111111111111011101011111111110111111101111101111111111100 \\
û & 1000101100110001 & E & 11111111111111111111111011101011111111110111111101111101111111111101000 \\
ø & 10001011001100101011 & È & 11111111111111111111111011101011111111110111111101111101111111111101010 \\
Ò & 10001011001100101101 & É & 11111111111111111111111011101011111111110111111101111101111111111101011 \\
ä & 1000101100110010111 & Æ & 1111111111111111111111101110101111111111011111110111110111111111110110 \\
É & 1000101100111 & Ö & 111111111111111111111110111010111111111101111111011111011111111111011110100 \\
B & 1001001 & Ú & 111111111111111111111110111010111111111101111111011111011111111111011110110 \\
R & 10011 & Ü & 11111111111111111111111011101011111111110111111101111101111111111101111011100 \\
ý & 11000000100 & U & 11111111111111111111111011101011111111110111111101111101111111111101111011110 \\
ì & 110000001010001 & J & 111111111111111111111110111010111111111101111111011111011111111111011111100 \\
â & 11000000101001 & Y & 111111111111111111111110111010111111111101111111011111011111111111011111110 \\
ï & 11000000101111 & Î & 111111111111111111111110111010111111111101111111011111011111111111011111111110 \\
J & 11000001 & Ò & 1111111111111111111111101110101111111111011111111110 \\
V & 1100001 & Ó & 11111111111111111111111011101011111111110111111101111111111000 \\
F & 11000110 & Ô & 11111111111111111111111011101011111111110111111101111111111001 \\
ã & 110001110100 & O & 11111111111111111111111011101011111111110111111101111111111010 \\
à & 1100011110 & ó & 11111111111111111111111011101011111111110111111101111111111110 \\
S & 11001 & ô & 11111111111111111111111011101011111111110111111101111111111111 \\
M & 110100 & è & 1111111111111111111111101110101111111111011111111000 \\
N & 11011 & é & 11111111111111111111111011101011111111110111111110010 \\
O & 11100 & û & 11111111111111111111111011101011111111110111111111010 \\
C & 111010 & Í & 111111111111111111111110111010111111111101111111111110 \\
À & 111011010000100 & I & 1111111111111111111111101110101111111111011111111111110 \\
ô & 11101101000011 & ß & 111111111111111111111110111010111111111110 \\
ç & 11101101001 & Â & 11111111111111111111111011101011111111111110 \\
Ú & 11110100001010000 & Û & 111111111111111111111110111010111111111111110 \\
Ø & 1111010000101000100001 & Ñ & 11111111111111111111111011101011111111111111110 \\
ß & 111101000010100010010 & F & 111111111111111111111110111011111111110010 \\
Ê & 111101000010100010011 & B & 1111111111111111111111101110111111111101000 \\
Å & 11110100001010001101000 & P & 11111111111111111111111011101111111111011000 \\
å & 111101000010100011011 & N & 11111111111111111111111011101111111111011100 \\
Ó & 1111010000101000111 & S & 11111111111111111111111011101111111111011110 \\
Í & 111101000010100111 & W & 1111111111111111111111101110111111111101111100 \\
Ç & 1111010000111111 & V & 11111111111111111111111011101111111111011111010 \\
í & 111101001 & G & 1111111111111111111111101110111111111101111110110 \\
I & 11111 & Z & 11111111111111111111111011101111111111011111110 \\

\hline
\end{tabular}
}\\
\end{center}

\vspace{-20pt}
\end{table*}

\vfill

\end{document}